%% file: main_accepted.tex
\documentclass[10pt]{article} 
\usepackage[accepted]{tmlr}

\input{math_commands.tex}

\usepackage{booktabs} 
\usepackage{hyperref}

\usepackage{amsmath}
\usepackage{amssymb}
\usepackage{mathtools}
\usepackage{amsthm}

\usepackage{bm}
\usepackage[capitalize,noabbrev]{cleveref}
\usepackage[table]{xcolor}
\usepackage{colortbl}
\usepackage{url}
\usepackage{ragged2e}
\usepackage{enumitem}
\usepackage{wrapfig}
\usepackage{array}
\usepackage{appendix}
\usepackage{amsthm}
\usepackage{graphicx} 
\usepackage{subcaption}
\usepackage{multirow}

\usepackage{longtable}
\usepackage{booktabs}
\usepackage{array}

\let\TMLRauthorAND\AND
\let\AND\relax

\usepackage{algorithm}
\usepackage{algorithmic}

\newcommand{\x}{x}
\newcommand{\y}{y}

\theoremstyle{plain}

\theoremstyle{definition}

\theoremstyle{remark}

\title{Inverse Problems Conditioned on Observation Ensembles: Applications and Methods}

\author{\name Zhengyan Huan \email  Zhengyan.Huan@tufts.edu \\
      \addr Department of Electrical and Computer Engineering, Tufts University\\
      NSF AI Institute for Artificial Intelligence and Fundamental
Interactions
      \AND
      \name Camila Pazos
 \email  Camila.Pazos@tufts.edu \\
      \addr  Department of Physics and Astronomy, Tufts University
      \AND
      \name
Martin Klassen
 \email  martin.klassen@cern.ch \\
      \addr  European Organization for Nuclear Research
      \AND
      \name Vincent Croft  \email  vincent.croft@cern.ch\\
      \addr Leiden Institute for Advanced Computer Science LIACS, Leiden University
      \AND
      \name Pierre-Hugues Beauchemin  \email  Hugo.Beauchemin@tufts.edu\\
      \addr Department of Physics and Astronomy, Tufts University\\NSF AI Institute for Artificial Intelligence and Fundamental Interactions
      \AND
      \name Shuchin Aeron  \email  shuchin@ece.tufts.edu\\
      \addr Department of Electrical and Computer Engineering, Tufts University\\
      NSF AI Institute for Artificial Intelligence and Fundamental
Interactions}

\def\month{09}  
\def\year{2026} 
\def\openreview{\url{https://openreview.net/forum?id=XG3BcxSex0}} 

\begin{document}

\let\AND\TMLRauthorAND
\maketitle
\let\AND\relax

\begin{abstract}
    We introduce a new multivariate statistical problem that we refer to as the Ensemble-conditioned Inverse Problem (EIP). The aim of EIP is to invert for an ensemble that is distributed according to the pushforward of a prior under a forward process. In high energy physics (HEP), this is related to a widely known problem called unfolding, which aims to reconstruct the true physics distribution from observations that are distorted by detector effects. The EIP also arises in full waveform inversion (FWI) and inverse imaging with unknown priors. We propose non-iterative inference-time methods that construct posterior samplers based on a new class of conditional generative models, which we call ensemble inverse generative models. For the posterior modeling, these models additionally use the ensemble information contained in the observation set on top of single observations.  Unlike existing methods, our proposed methods avoid explicit and iterative use of the forward model at inference time via training across several sets of truth-observation pairs that are consistent with the same forward model, but originate from a wide range of priors. We empirically demonstrate that this training procedure can implicitly encode the likelihood model, enabling direct posterior inference for unseen priors to some degree. We benchmark the proposed method on several synthetic and real datasets in inverse imaging, HEP, and FWI. Our code is available at \url{https://github.com/ZhengyanHuan/EIP}.
\end{abstract}

\section{Introduction}
In a typical inverse problem, one is interested in recovering a ground-truth signal from an observation using a likelihood or forward model. The true signal is usually assumed to follow a prior distribution, which can enforce regularity when the forward model makes the problem ill-posed. In this paper, we are motivated by applications where one has access to multiple datasets, each consisting of truth–observation pairs generated under a different prior but the same forward model. However, the forward model is either unknown or too computationally expensive to use at inference time. We call such a problem an observation-ensemble-conditioned inverse problem when the recovery of either the latent distribution or a sample-wise posterior is conditioned on an ensemble of observations whose elements share an unknown underlying prior. In other words, the goal is to use the training datasets to map a new observation distribution to its corresponding prior and/or posteriors without explicitly using the forward model. For brevity, we refer to this setting as the Ensemble-conditioned Inverse Problem (EIP) throughout the paper.

\begin{figure}
    \centering
    \raisebox{-0.5\height}{
        \includegraphics[width=0.4\linewidth]{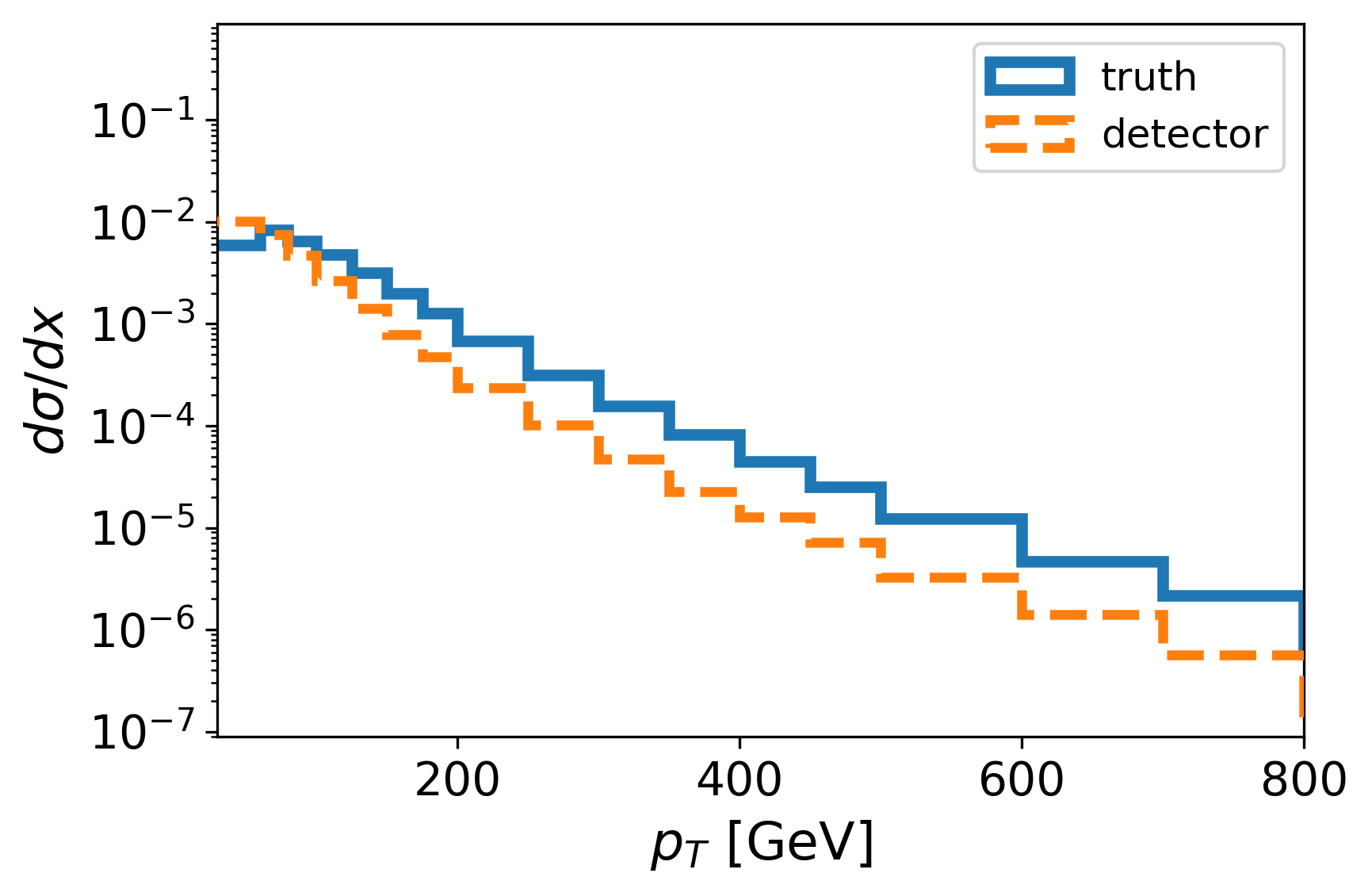}
    }
    \hfill
    \raisebox{-0.5\height}{
        \includegraphics[width=0.55\linewidth]{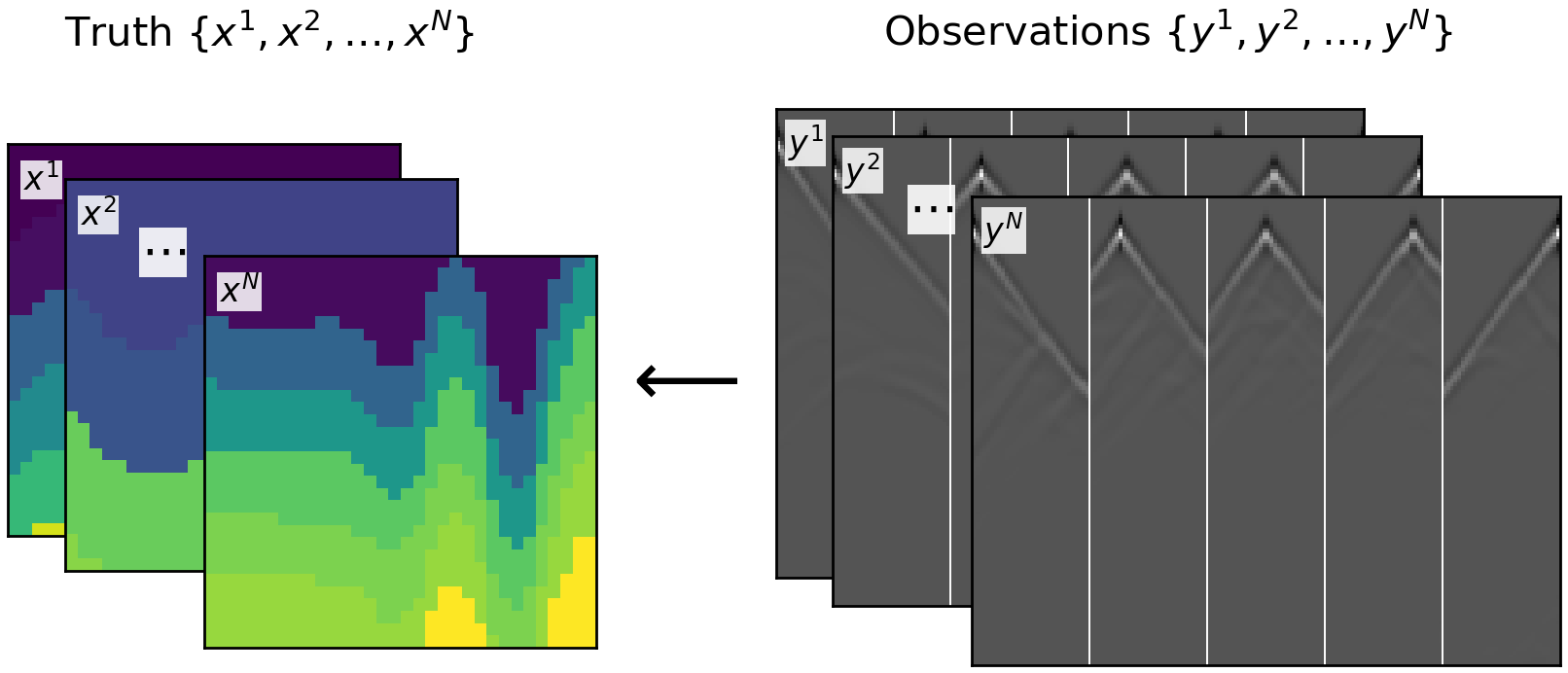}
    }
    \caption{Examples of EIPs. Left: HEP unfolding, where the truth-level prior distribution is recovered from the noisy detector-level distribution. Right: seismic full waveform inversion, where a subsurface velocity map from an unknown prior is reconstructed from the corresponding seismic waveform observations.}
    \label{fig:EIPexamples}
    \vspace{-10pt}
\end{figure}

An important application of EIP arises in the high-energy physics (HEP) domain, where one \emph{unfolds} to remove detector effects \citep{ibu_dagostini, OmniFold}, as illustrated in the left panel of Fig.~\ref{fig:EIPexamples}. In this example, we are given detector-level samples that form the dashed-orange histogram, and the task is to recover the prior, denoted in the solid blue histogram.
Another domain where EIP arises naturally is the seismic full waveform inversion (FWI) problem \citep{NEURIPS2022_FWI}, where one wants to recover the subsurface velocity maps from the received seismic data with an unknown prior, as illustrated in the right panel of Fig.~\ref{fig:EIPexamples}. In FWI the forward models are typically very expensive to simulate or are simply not given.
EIP also arises in inverse imaging, where one wants to de-corrupt a set of corrupted images with an unknown prior \citep{Ambient, hu2023restoration}. A set of recent works has considered EIP-like problems arising in contexts of large language models and in-context learning \citep{m2m,teh2025solving,adu2024approximate}. In \citet{m2m,adu2024approximate} the main problem is to understand if given pairs of measures, whether there exists a transformer architecture that can map a given input to its corresponding output, thus learning a measure to measure map. 
This is related to EIP’s objective of learning a distribution-dependent mapping from an observation measure to its associated prior or posterior. The setting of \citet{teh2025solving} also comes close to EIP. \citet{teh2025solving} proposes to use a transformer to infer the hidden parameters in a Poisson forward process, provided with a set of observations. 

To formalize EIP, let $\x \in \mathbb{R}^d$ be a random variable with a prior distribution $p(x)$. We consider inverse problems in which the observation $y$ is related to the unknown 
$x$ through an additive-noise \emph{{forward model}}: 
\begin{equation}
\label{eq:forward_model}
    \y = F(\x)+n(\x), \tag{Fwd-Model}
\end{equation}
where $F$ is a forward (measurement) operator and $n(\x)$ represents an additive noise, which can in general depend on $\x$. 
Within this setup, we consider the following problem that we refer to as the Ensemble-conditioned Inverse Problem (EIP). We are given a dataset $\mathcal{D}=\{\mathcal{D}_1,\cdots, \mathcal{D}_M\}$ consisting of multiple truth-observation pairs arising from sampling observations via \eqref{eq:forward_model} from $M$ prior distributions $p_{m}, m \in [1:M]$. 
\begin{equation}
    \mathcal{D}_m = \{(\x^{m,j},\y^{m,j})\}_{j=1}^{N_m} \overset{i.i.d.}{\sim} p_m(x)p(y|x), 
\end{equation}
where $(\x^{m,j},\y^{m,j})$ denotes the $j$-th truth-observation pair in $\mathcal{D}_m$, and the size of $\mathcal{D}_m$ is $N_m$. The pairs $(\x^{m,j},\y^{m,j})$ are independently and identically distributed (i.i.d.) according to the joint distribution $p_m(x)p(y|x)$, and the conditional distribution $p(y|x)$ is determined via \eqref{eq:forward_model} and is the same for all datasets $\{\mathcal{D}_1,\cdots, \mathcal{D}_M\}$. We assume that we only have access to $\mathcal{D}$ and no direct knowledge about \eqref{eq:forward_model}. 

\vspace{-5pt}
\paragraph{Problem statement (EIP-I for the prior):} Given training data $\mathcal{D}$, and given a new set of observations $\mathcal{Y} =\{ y^1, \cdots, y^{N}\}$ obtained from an unknown prior $p(x)$ and the same (as $\mathcal{D}$, but unknown) forward model, generate samples $x^1, \cdots, x^{N_{\mathrm{gen}}} | \mathcal{Y}$ such that for a given $\lambda > 0 $, 
\begin{equation}
    \rho(\hat{p}(x|\mathcal{Y}),p(x)) < \lambda,
\end{equation}
where $\hat{p}(x|\mathcal{Y}) =\lim_{N_{\mathrm{gen}}\rightarrow\infty}\frac{1}{N_{\mathrm{gen}}} \sum_{n=1}^{N_{\mathrm{gen}}} \delta_{x^n|\mathcal{Y}}$ is the limiting empirical measure corresponding to the generated samples.
$\rho(\cdot,\cdot)$ denotes a discrepancy measure between distributions, such as the Kullback–Leibler divergence, total variation \citep{KL}, or Wasserstein distance \citep{WD1}, and the Dirac delta function $\delta_{x^n}$ denotes the probability density of a distribution concentrated at the $n$-th generated sample $x^n$.
In other words, the aim of EIP-I is to generate samples whose distribution comes close to the prior distribution that leads to the observations. 
For practical utility that will become clear in the exposition later, we restrict the EIP-I problem further to learn to generate samples via posterior sampling, given observations from a prior.

\paragraph{Problem statement (EIP-II for the posterior):} Given training data $\mathcal{D}$, and given a new set of $i.i.d.$ observations $\mathcal{Y} =\{ y^1, \cdots, y^{N}\}$ obtained from an unknown prior $p(x)$ and the same (as $\mathcal{D}$) but unknown forward model, for any given $y$, generate conditional samples {$x^1, \cdots, x^{N_{\mathrm{gen}}} | y, \mathcal{Y} $} such that for a given $\lambda > 0 $, 
\begin{equation}
  \rho(\hat{p}(x|y,\mathcal{Y}),p(x|y)) < \lambda, 
\end{equation}
where $\hat{p}(x|y,\mathcal{Y}) =\lim_{N_{\mathrm{gen}}\rightarrow\infty}\frac{1}{N_{\mathrm{gen}}} \sum_{n=1}^{N_{\mathrm{gen}}} \delta_{x^n|y,\mathcal{Y}}$, and $p(x|y) = \frac{p(x) p(y|x)}{p(y)}$. In EIP, it is assumed that the observation set $\mathcal{Y}$ is sufficiently informative, in the sense that it contains enough distributional information to aid prior/posterior inference. More precisely, $\mathcal{Y}$'s informativeness depends on whether the mapping from a candidate prior to its induced observation distribution can distinguish among the priors under consideration. When the induced observation distributions are distinct but weakly separated, a larger set size $N$ is required for the empirical distribution represented by $\mathcal{Y}$ to reliably capture their differences. We discuss the practical effect and selection of $N$ further in Sec.~\ref{sec:appNN}.

It is evident that the integration of the solution to EIP-II yields a good approximation of the solution to EIP-I. However, the integration of posteriors that are not the solution to EIP-II can still be the solution to EIP-I. We refer the readers to the Gaussian example in Sec.~3 in \citet{butter2025generative} and our example in Fig.~\ref{fig:introf} in Appendix~\ref{app:eip_relationship}.

\begin{table*}[htb]
\centering
\resizebox{\textwidth}{!}
{
\begin{tabular}{|c|c|c|c|c|c|c|}
\hline
 & \multicolumn{1}{c|}{Method} & Requirements & Objective &Iterative & \begin{tabular}[c]{@{}c@{}}Tunable\\ Regularization\end{tabular} & \begin{tabular}[c]{@{}c@{}}Designed to recover\\ unseen priors\end{tabular}  \\ \hline
\multirow{3}{*}{Non-ML} & IBU  & \eqref{eq:forward_model} & $p(x)$ & Yes & Yes & Yes \\ \cline{2-7} 
 & SVD Unfolding & \eqref{eq:forward_model} & $p(x)$ &Partial& Yes & Yes \\ \cline{2-7} 
  & Measure decomposition & \eqref{eq:forward_model} & $p(x|y)$ &Yes& Yes & Yes  \\ \hline
 \multirow{1}{*}{Theoretical} &Measure-to-measure interpolation    & $\mathcal{D}$ & $p(x)$&No & No & No  \\ \hline
\multirow{10}{*}{ML-based}
& OmniFold  & $\mathcal{D}$ or \eqref{eq:forward_model} & $p(x)$&Yes & Yes & Yes \\ \cline{2-7} 
 & GANs & \eqref{eq:forward_model} & $p(x)$ &No& No & Yes \\ \cline{2-7} 
  & DPnP & $\{x^j\}_{j=1}^N$ and \eqref{eq:forward_model} & $p(x|y)$ &No& No & No \\ \cline{2-7} 
 & Ambient diffusion  & $\{y^{j}\}_{j=1}^{N}$ and \eqref{eq:forward_model} & $p(x|y)$&No & No & No \\ \cline{2-7} 
   & IcINN  & \eqref{eq:forward_model} & $p(x|y)$&Yes & No & Yes \\ \cline{2-7} 
 & SBUnfold & $\mathcal{D}$ & $p(x|y)$&No & No & Yes \\ \cline{2-7} 
  & DDRM  & \begin{tabular}[c]{@{}c@{}}Pretrained model and\\ \eqref{eq:forward_model}\end{tabular}  & $p(x|y)$& No & No & No \\ \cline{2-7} 
 & GDDPM & $\mathcal{D}$ & $p(x|y)$& No & No & Yes \\ \cline{2-7} 
 & Ours & $\mathcal{D}$ & $p(x|y)$&No & No & Yes  \\ \hline
\end{tabular}}
\caption{Comparison of methods for solving EIP-I (objective: $p(x)$) \& EIP-II (objective: $p(x|y)$) and their key characteristics.
Iterative Bayesian unfolding (IBU) appears in \citet{ibu_dagostini}. 
Singular value decomposition (SVD) Unfolding appears in \citet{SVDUnfold}.
Measure decomposition method for posterior sampling appears in \citet{Andrea2025Provably}.
Measure-to-measure interpolation approaches appear in \citet{m2m, adu2024approximate}. 
 OmniFold appears in \citet{OmniFold}.
 Generative adversarial networks (GANs) for inverse problems appear in \citet{GAN_away_2020, GANunfold2018}.  Diffusion plug-and-play (DPnP) method appears in \citet{DPnP}. 
 Ambient diffusion appears in \citet{Ambient}. Iterative conditional invertible neural networks (IcINN) approaches appear in \citet{cINNunfold2024}.
 SBUnfold appears in \citet{diefenbacher2023improving}.
 Denoising diffusion restoration model (DDRM) appears in \citet{DDRM}.
  Generalizable conditional denoising diffusion probabilistic model (GDDPM) appears in  \citet{Camila24}. ``iterative'' refers to inference-time refinement of an estimated
latent distribution using the forward model or paired simulation data to match the observed distribution.
}
\vspace{-10pt}
\label{tb:comparison}
\end{table*}
\subsection{Related Works}
Table~\ref{tb:comparison} provides a summary of key features among non-ML, theoretical, and ML-based methods for solving the EIP and / or classical inverse problem. 
\vspace{-5pt}
\begin{enumerate}[leftmargin=1em]
\item \textbf{Non-ML methods:}
Traditional methods designed for unfolding reconstruct the prior via iterative probabilistic updates IBU \citep{ibu_dagostini} and suppression of contributions with small singular values \citep{SVDUnfold}.  Both methods require explicit modeling of the forward process and binned data. In a more general setting, \citet{Andrea2025Provably} proposes an iterative posterior measure decomposition method that enables efficient sampling for sparse Bayesian inverse problems.
\vspace{-5pt}
\item 
\textbf{Theoretical methods:} \citet{m2m,adu2024approximate} provide mathematical frameworks for understanding transformers as measure-to-measure maps and prove that a single transformer can approximate the transport maps between multiple distribution pairs. The depth and complexity of the transformer depend on the structure and the number of pairs. However, the problem of generalization to unseen measures was not considered, and no algorithm was proposed for solving the EIP. \citet{teh2025solving} proves that transformers can approximate classical empirical Bayes estimators and proposes a training algorithm. Nevertheless, this method is limited to the one-dimensional Poisson–EB setting.
\vspace{-5pt}
\item  \textbf{ML-based methods:} OmniFold \citep{OmniFold} is a representative iterative re-weighting method for unfolding that shapes a given prior to the target prior. Generative methods have also become successful tools for addressing inverse problems, leading to a surge of approaches, including GANs \citep{GAN_away_2020,GANunfold2018}, DPnP \citep{DPnP}, ambient diffusion, and SBUnfold \citep{diefenbacher2023improving}.
In particular, GDDPM \citep{Camila24} aims to solve EIP-II via posterior modeling and sampling. Built based on conditional DDPM (cDDPM), GDDPM additionally utilizes moment information of observations to support generalization ability across different physics processes. With the objective of avoiding computationally costly iterative inference, bypassing the difficulty of obtaining the forward model, and effectively incorporating distributional information embedded in observations, this work provides a framework for solving EIP-II via generative models. 
\end{enumerate}

\vspace{-10pt}
\subsection{Contributions}
We list the contributions of this work as follows:
\begin{enumerate}[leftmargin=1em]
    \item This work proposes a novel non-iterative framework for solving EIP-II, called ensemble inverse generative models, which models the posterior and conditions sampling on both individual observations and their corresponding observation sets.
    \vspace{-5pt}
    \item With the ensemble information extracted via a permutation-invariant structure from the observation set, the proposed method learns a distribution-dependent posterior sampler,
    demonstrating a superior posterior inference ability and a strong generalization ability to unseen priors. \vspace{-5pt}
    \item Under several synthetic settings and real applications, including HEP unfolding, FWI, and image inversion tasks,  we demonstrate that the proposed methods outperform baselines without relying on explicit knowledge about the priors and the forward model.
\end{enumerate}

\vspace{-10pt}
\section{Method}
We address EIP-II via a non-iterative posterior sampling method. Specifically, generative models that are conditioned on not only the single observation $y$ but also the observation set $\mathcal{Y}$
are utilized to model the posterior and serve as a posterior sampler. With the aid of ensemble information extracted from the observation set $\mathcal{Y}$, the proposed method is shown to have a strong inductive bias to unseen priors. To state the methods, we refer the readers to two successful generative models, viz., Denoising Diffusion Probabilistic Models (DDPM) \citep{NEURIPS2020_DDPM} and Flow Matching (FM) \citep{lipman2023flow} for background, and we provide more details to the conditional versions of them in Sec.~\ref{sec:introDDPMFM}. The main notations used in the proposed framework are summarized in Appendix~\ref{app:illustration_notation} Table~\ref{tab:notation}.

\subsection{Ensemble Inverse Generative Models for EIP-II}


Our main idea behind addressing EIP-II is that the observation set $\mathcal{Y}$, in which all observations yield from a single prior distribution $p(x)$, contains information about $p(x)$. 
This prior information is not directly available, but can contribute towards a valid posterior inference for any given $y$ yielding from $p(x)$. Inspired by \citet{teh2025solving, Camila24} and with the objective of utilizing the ensemble information contained in $\mathcal{Y}$,
our recovery model is conditioned on not only the observation $y$ but also the observation set $\mathcal{Y}$. The size of $\mathcal{Y}$ should be large enough to reflect the underlying ensemble information. However, in conditional generative modeling, directly conditioning on a large input set can be computationally inefficient and statistically unstable, as the model must process high-dimensional and unordered data. To address this, one can first encode the set using a \emph{permutation invariant} structure, such as using the moment function as in \citet{Camila24}. For a more versatile and adaptive representation, we propose to extract the ensemble information via $\phi_w:\mathbb{R}^{N\times d}\rightarrow\mathbb{R}^{k}$, a permutation invariant neural network (NN) parameterized with $w$, that maps an observation set $\mathcal{Y}$ containing $N$ $d$-dimensional samples into a $k$-dimensional representation that reflects the ensemble information. 
Formally, let $S_{N}$ denote the set of all permutations of indices $\{1,2,\cdots,N\}$. $\phi_w$ should satisfy
\begin{equation}
    \forall s \in S_{N}, ~~\phi_w(s \mathcal{Y}) = \phi_w(\mathcal{Y}),~~\mathcal{Y} = \{y^1,\cdots,y^{N}\}.
\end{equation}
This allows $\phi_w$ to process $\mathcal{Y}$ as a set, focusing on the group feature and ignoring the order information.
 Choices for implementing $\phi_w$ include deep set \citep{Deepset} and set transformer \citep{settransformer}. 

Based on this insight, we propose an algorithm for solving EIP-II, named ensemble inverse denoising diffusion probabilistic model (EI-DDPM) / ensemble inverse flow matching (EI-FM), as presented in Alg.~\ref{alg:train} and Alg.~\ref{alg:sample_imp}. 
EI-DDPM / EI-FM is based on conditional-DDPM / conditional-FM frameworks, wherein an NN denoted by $\varepsilon_\theta$, parameterized by $\theta$ is employed to predict the noise / velocity field at each step. In addition to the intermediate states $x_t$ and time information $t$, $\varepsilon_\theta$ accepts single observations $y$, as well as the ensemble information $\phi_w(\mathcal{Y})$ as inputs in order to model the posterior $p(x|y,\mathcal{Y})$ in EIP-II. The observation $y$ is contained in the observation set $\mathcal{Y}$. $y$ directly specifies the observed sample to be inverted. $\phi_w(\mathcal{Y})$ serves as the ensemble information that facilitates the posterior inference for the observation $y$.
Although the dimension of the ensemble information $k$ is determined by the user, we generally set it to the same order as $d$ because $y\in \mathbb{R}^d$ and $\phi_w(\mathcal{Y})\in \mathbb{R}^k$ are jointly provided as conditioning inputs. Comparable dimensions give them similar representational capacity and avoid an over-restricted ensemble representation. Provided with truth-observation pairs resulting from sufficiently diverse priors, $\varepsilon_\theta$ and $\phi_w$ combined is able to generalize for posteriors induced by previously unseen priors. We numerically illustrate these features in 
Sec.~\ref{sec:exp}.

The stability of the learned representation of ensemble information $\phi_w(\mathcal{Y})$ depends on an extra hyperparameter $N$ -- the number of samples in $\mathcal{Y}$. First, $N$ should be large enough for $\mathcal{Y}$ to have the capability to represent the distributional information of $p(y)$, thus being able to contain valid ensemble information. Second, considering that $N$ is fixed during the training stage in Alg.~\ref{alg:train}, the input observation set size for Alg.~\ref{alg:sample_imp} of inference should remain $N$ for robustness. Therefore, it is important to discuss cases in which the available observation set size $N' \neq N$, at inference time.
For the case $N'>N$, 
subsets of size $N$ can be picked repeatedly to perform Alg.~\ref{alg:sample_imp} until the union of the subsets fully covers the target observation set. For the case $N'<N$, one can randomly duplicate $N-N'$ samples so that the set size is expanded to $N$. 
For target sets with $N'\ll N$, Alg.~\ref{alg:sample_imp} with duplication strategy may perform poorly since a set with too many duplicates will reflect highly incorrect ensemble information. The effects of $N$ and $N'$ are further discussed and numerically investigated in Sec.~\ref{sec:appNN}.

The above construction can be viewed as a distribution-level analogue of supervised regression. In standard regression, a model generalizes by learning a stable sample-to-sample relationship that holds across samples drawn from the same underlying distribution, rather than by memorizing individual inputs. In EIP-II, however, the posterior is generally not determined by a single observation $y$ alone; it also depends on the unknown prior that generates the observation set. A vanilla conditional generative model conditioned only on $y$ therefore learns a sample-wise posterior averaged over the training priors, which may limit its ability to adapt to unseen priors. By contrast, EI-DDPM and EI-FM additionally condition on $\phi_w(\mathcal{Y})$, which serves as a learned summary of the empirical observation distribution. This turns the posterior sampler from a sample-wise map $p(x|y)$ into a distribution-dependent map $p(x|y,\mathcal{Y})$. Compared with GDDPM, which conditions a DDPM on predefined moment statistics of physics-particle observables, our framework jointly learns the ensemble representation $\phi_w$ and the posterior sampler $\varepsilon_\theta$ through the generative objective. The learned representation can capture nonlinear and cross-variable dependencies that may be missed by prescribed low-order moments. Moreover, the proposed framework requires no domain-specific selection of summary statistics. Consequently, the ensemble inverse models learn a stable relationship from an observation ensemble to its induced posterior distribution, analogous to the stable sample-to-sample relationship learned in regression. This provides the proposed methods with an inductive bias for posterior inference and generalization to unseen priors.

\begin{algorithm}[H]
\caption{EI-DDPM's and EI-FM's Training algorithm}
\label{alg:train}
\begin{algorithmic}
\STATE \textbf{Input:}
EI-DDPM/FM's neural network $\varepsilon_\theta$, ensemble information neural network $\phi_w$, observation set size $N$, training datasets $\mathcal{D}=\{\mathcal{D}_1,\cdots,\mathcal{D}_M\}$, EI-DDPM's schedule parameters \{$\beta_t, \alpha_t,\bar{\alpha}_t, T$\}, learning rate $\eta$
\STATE \textbf{Output:} Trained $\varepsilon_\theta, \phi_w$
\STATE ~

\REPEAT
    \STATE Choose $m\sim \text{Uniform}(\{1,\cdots,M\})$ and draw a subset of $N$ pairs
    $\{(\x^{m,j},\y^{m,j})\}_{j=1}^{N}$ from $\mathcal{D}_m$
    \STATE $\mathcal{Y}\gets \{y^{m,j}\}_{j=1}^N$, $\mathcal{L}(\theta,w)\gets 0$
    \FOR{each $(x,y)$ pair in the subset}
        \STATE
        \begin{minipage}[t]{0.48\linewidth}
            \textbf{if} using EI-DDPM \textbf{then}\\
            \hspace*{1em}$t\sim \text{Uniform}(\{1,\cdots,T\})$, $\xi\sim\mathcal{N}(\bm{0},\bm{I})$\\
            \hspace*{1em}$\x_t\gets\sqrt{\bar{\alpha}_t}\x
            +\sqrt{1-\bar{\alpha}_t}\xi$\\
            \hspace*{1em}$u\gets\xi$\\
            \textbf{end if}
        \end{minipage}\hfill
        \begin{minipage}[t]{0.48\linewidth}
            \textbf{if} using EI-FM \textbf{then}\\
            \hspace*{1em}$t\sim\mathcal{U}[0,1]$, $\xi\sim\mathcal{N}(\bm{0},\bm{I})$\\
            \hspace*{1em}$\x_t\gets t\x+(1-t)\xi$\\
            \hspace*{1em}$u\gets\x-\xi$\\
            \textbf{end if}
        \end{minipage}
        \STATE $\mathcal{L}(\theta,w)\gets
        \mathcal{L}(\theta,w)+
        \frac{1}{N}\left\|
        \varepsilon_\theta(\x_t,t,\y,\phi_w(\mathcal{Y}))-u
        \right\|_2^2$
    \ENDFOR
    \STATE $(\theta,w)\gets
    (\theta,w)-\eta\nabla\mathcal{L}(\theta,w)$
\UNTIL{converged}

\STATE \textbf{Return} $\varepsilon_\theta, \phi_w$
\end{algorithmic}
\end{algorithm}

\vspace{-10pt}
\begin{algorithm}[H]
\caption{EI-DDPM's and EI-FM's sampling algorithm}
\label{alg:sample_imp}

\begin{algorithmic}
\STATE \textbf{Input:}
EI-DDPM/FM's neural network $\varepsilon_\theta$, ensemble information neural network $\phi_w$, observation set $\mathcal{Y}=\{y^{j}\}_{j=1}^N$, EI-DDPM's schedule parameters \{$\alpha_t,\bar{\alpha}_t, \sigma_t,T$\}, EI-FM's discretization interval $\Delta t$
\STATE \textbf{Output:} Recovery set $\{\hat{x}^{j}\}_{j=1}^N$
\STATE ~
\STATE $z = \phi_w(\mathcal{Y})$
\end{algorithmic}

\noindent
\begin{minipage}[t]{0.49\linewidth}
\begin{algorithmic}
\IF{using EI-DDPM}
    \FOR{$j=1,2,\cdots,N$}
        \STATE $\x_T\sim\mathcal{N}(\bm{0},\bm{I})$
        \FOR{$t=T\cdots,1$}
            \STATE $\xi\sim\mathcal{N}(\bm{0},\bm{I})$
            if $t>1$, else $\xi\gets 0$
            \STATE $e\gets\varepsilon_\theta(\x_t,t,y^{j},z)$
            \STATE $\displaystyle
            \x_{t-1}\gets
            \frac{1}{\sqrt{\alpha_t}}
            \left(\x_t-
            \frac{1-\alpha_t}{\sqrt{1-\bar{\alpha}_t}}e\right)
            +\sigma_t\xi$
        \ENDFOR
        \STATE $\hat{x}^{j}\gets\x_0$
    \ENDFOR
\ENDIF
\end{algorithmic}
\end{minipage}\hfill
\begin{minipage}[t]{0.49\linewidth}
\begin{algorithmic}
\IF{using EI-FM}
    \FOR{$j=1,2,\cdots,N$}
        \STATE $\x_0\sim\mathcal{N}(\bm{0},\bm{I}), t\gets 0$
        \REPEAT
            \STATE $\x_{t+\Delta t}\gets\x_t+
            \varepsilon_\theta(\x_t,t,y^{j},z)\Delta t$
            \STATE $t\gets t+\Delta t$
        \UNTIL{$t=1$}
        \STATE $\hat{x}^{j}\gets\x_1$
    \ENDFOR
\ENDIF
\end{algorithmic}
\end{minipage}

\begin{algorithmic}
\STATE \textbf{Return} $\{\hat{x}^{j}\}_{j=1}^N$
\end{algorithmic}
\end{algorithm}

\section{Experiments}
\label{sec:exp}
\subsection{Baselines for Comparison}
\paragraph{Conditional DDPM (cDDPM) and conditional FM (cFM):} cDDPM and cFM model the posterior $p(x|y)$ with the conditional variable incorporating only a single observation. No ensemble information is included.
\vspace{-3mm}
\paragraph{Generalizable DDPM (GDDPM) \citep{Camila24}:} GDDPM is built upon cDDPM and it incorporates additional moment information computed from $\mathcal{Y}$.
\vspace{-3mm}
\paragraph{OmniFold \citep{OmniFold}:}
OmniFold is a reweighting-based unfolding method that reweighs a given initial distribution towards the prior. The initial distribution is a critical factor in recovery performance. Since in the EIP setup, we are provided with $\{({x^{m,j},\y^{m,j})}\}_{j=1}^{N_m}, m= 1,\cdots,M$, we consider two ways of selecting the initial distribution to invert for a set of observations $\mathcal{Y}'$. a) \textbf{OmniFold-best:} Picking $m^\star$, such that $\{y^{m^\star,j}\}_{j=1}^{N_{m^\star}}$ has the minimum sliced Wasserstein distance (SWD)\footnote{SWD measures the similarity between two distributions, with smaller values indicating greater similarity.} \citep{SWD} from $\mathcal{Y'}$, and $\{x^{m^\star,j}\}_{j=1}^{N_{m^\star}}$ serves as the initial distribution; and b) \textbf{OmniFold-combine:} Using the mixture of all available priors $\{x^{m,j}\}_{j=1}^{N_{m}}, m=1,\cdots, M$ as the initial distribution.
\paragraph{SBUnfold \citep{diefenbacher2023improving}:} SBUnfold leverages Schrodinger bridges with diffusion models to map observations to their truth. 

\paragraph{Sourcerer \citep{vetter2024sourcerer}:} Sourcerer is a sample-based method for learning a maximum-entropy prior distribution whose simulated outputs minimize a sample-based discrepancy, e.g., SWD, from the observed data. It requires an available differentiable forward model or a differentiable surrogate of it. In our case, an available differentiable forward model is not directly accessible; however, a surrogate can be trained based on the truth-observation pairs.

The NN structures for cDDPM, cFM, SBUnfold, and $\varepsilon_\theta$ used in EI-DDPM / EI-FM are kept the same (with input dimensions adjusted to match their respective inputs) for a fair comparison. We use the set transformer \citep{settransformer} structure for the implementation of $\phi_w$.

\subsection{Synthetic EIP}\label{sec:gaussian}
 
We first present a toy example of inverting for a perturbed $2$-D Gaussian distribution to demonstrate the effectiveness of the proposed method. The prior is a bivariate Gaussian distribution with mean $[0,0]^\top$ and covariance matrix $\left[\begin{matrix}
    1&\gamma\\\gamma&1
\end{matrix}\right]$, where $\gamma\in [-1,1]$ represents the correlation coefficient between the two dimensions. Let $x=[x_1,x_2]^\top\in \mathbb{R}^2$ denote a sample from the prior. The prior is given as
\begin{equation}
    x|\gamma\sim \mathcal{N}\left(\left[\begin{matrix}
        0\\0
    \end{matrix}\right], \left[\begin{matrix}
    1&\gamma\\\gamma&1
\end{matrix}\right]\right).
\end{equation}
In this EIP, we consider that $x$ undergoes a linear transformation by a matrix $A\in \mathbb{R}^{2\times 2}$, and is perturbed by an additive noise term $n(x)\in \mathbb{R}^2$. The observed signal $y\in \mathbb{R}^2$ is given by
\begin{equation}\label{eq:toyforward}
\begin{aligned}
    y = Ax+n(x),\quad A = \left[\begin{matrix}
    1&0.5\\0.5&2
\end{matrix}\right],\quad n(x)\sim \mathcal{N}\left(\left[\begin{matrix}
    0.2x_1\\0.2x_2
\end{matrix}\right], \left[\begin{matrix}
    0.25\|x\|_2^2&0\\0&0.25\|x\|_2^2
\end{matrix}\right]\right).
\end{aligned}
\end{equation}
The objective is to recover the prior given its observation set $\mathcal{Y}$ corresponding to an unknown $\gamma$.

\begin{figure*}[htb]
    \centering
        \begin{tabular}{cccccc}
         \begin{subfigure}{0.19\textwidth}
            \includegraphics[width=\linewidth]{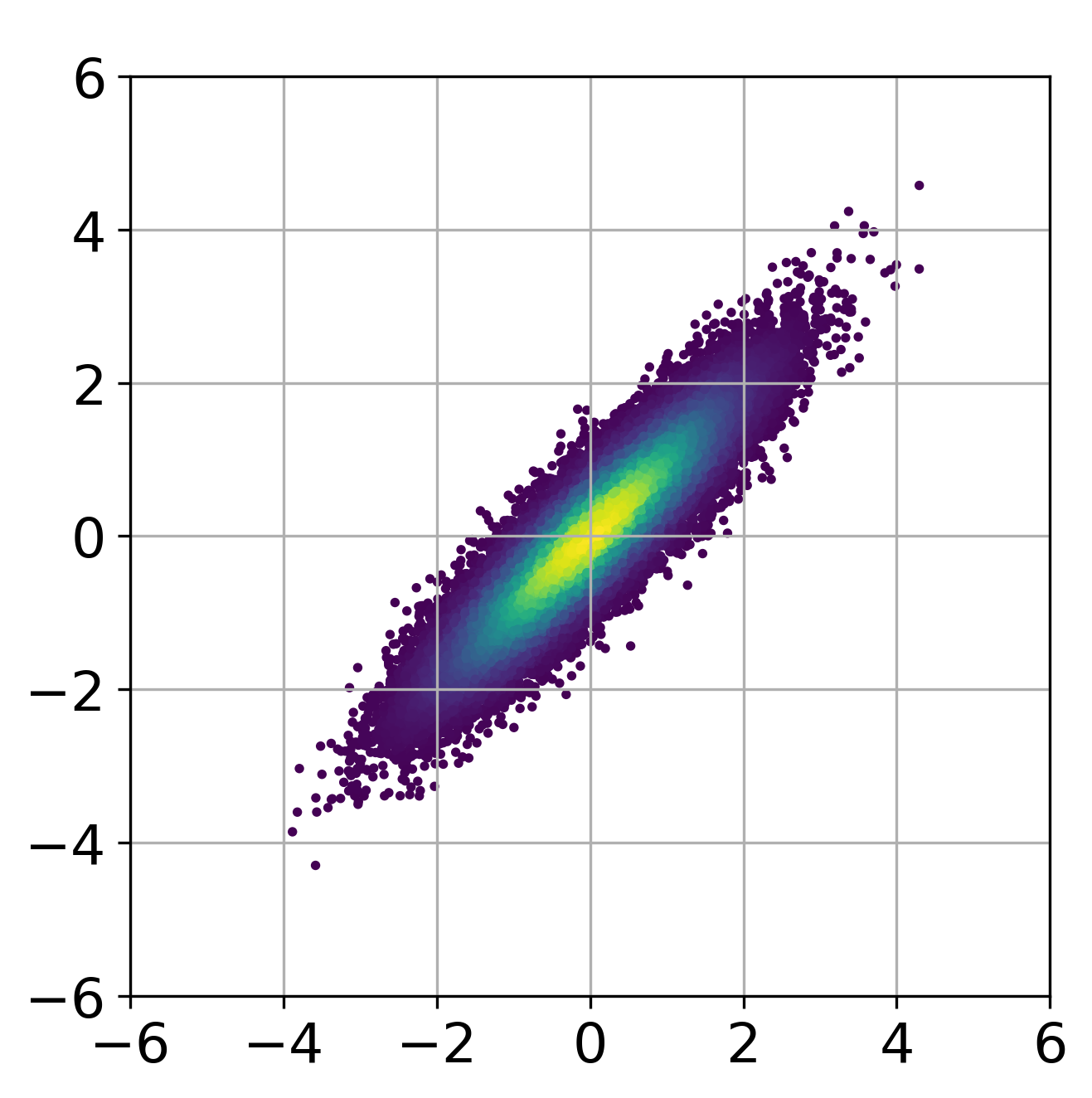}
            \caption{Prior }
        \end{subfigure} &\hspace{-15pt}
                \begin{subfigure}{0.19\textwidth}
            \includegraphics[width=\linewidth]{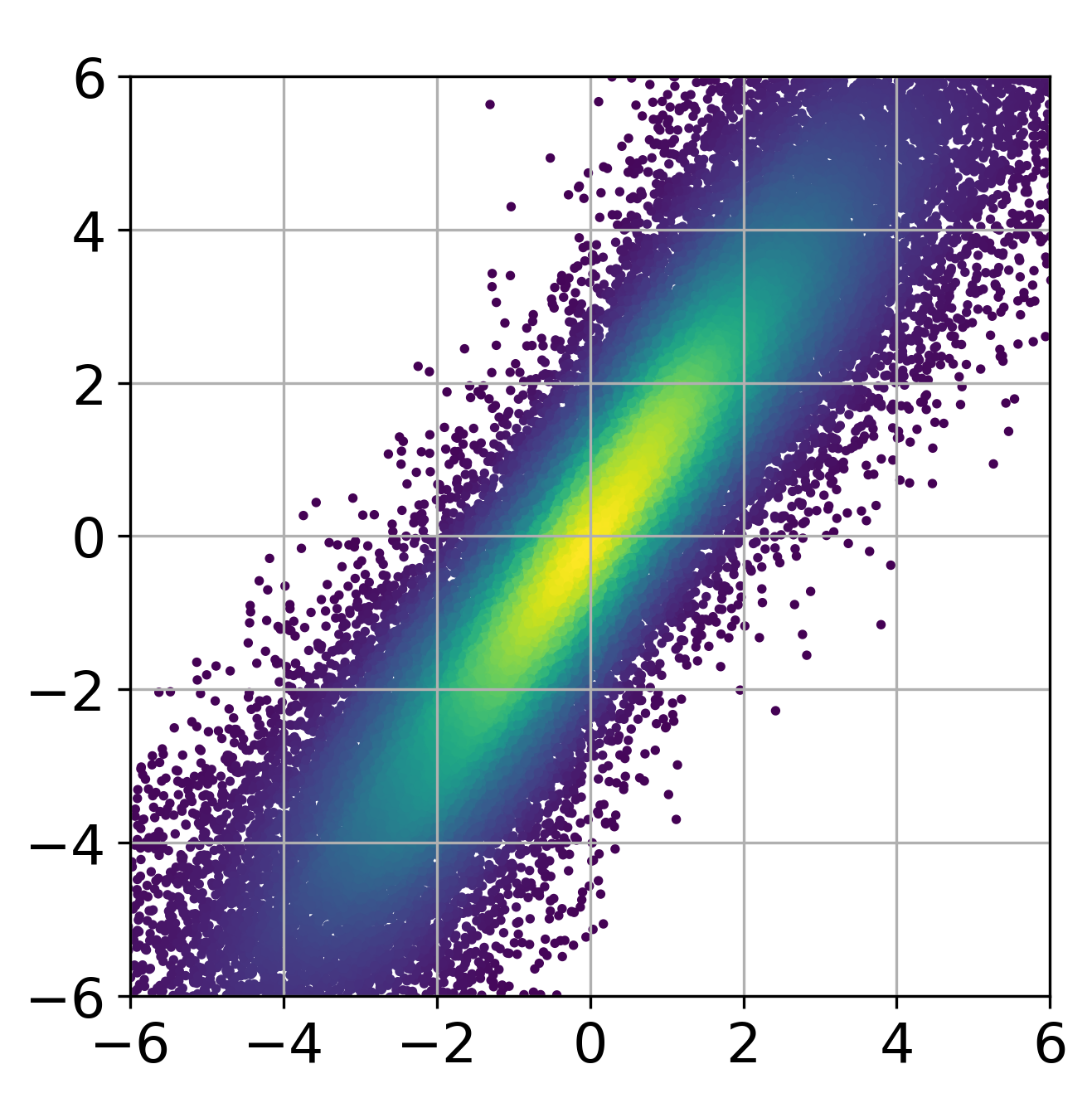}
            \caption{Observation}
        \end{subfigure} &\hspace{-15pt}
       
        \begin{subfigure}{0.19\textwidth}
            \includegraphics[width=\linewidth]{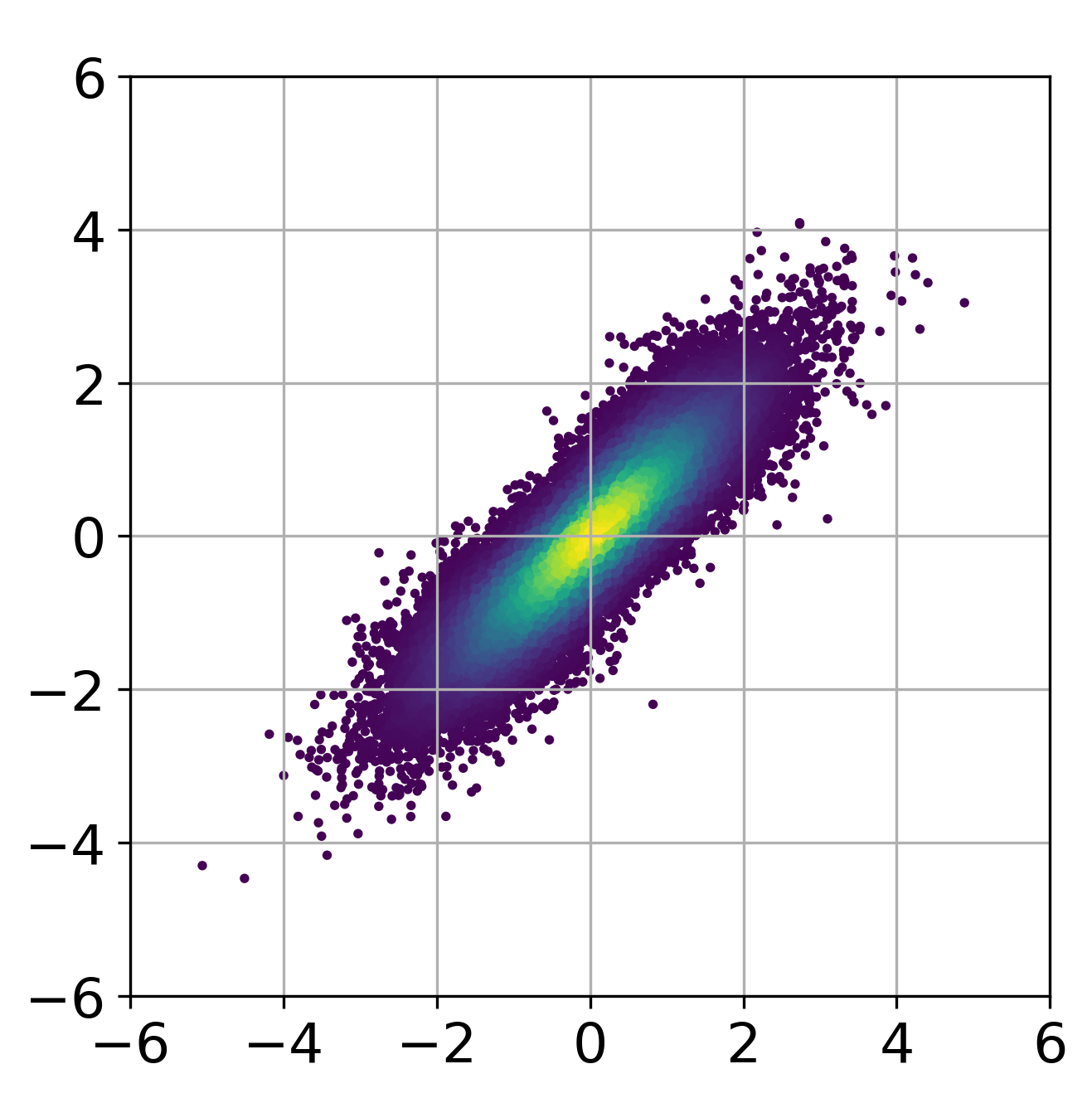}
            \caption{{\fontsize{6.5}{10}\selectfont EI-FM (SWD$=0.028$)} }
        \end{subfigure} &\hspace{-15pt}
        \begin{subfigure}{0.19\textwidth}
            \includegraphics[width=\linewidth]{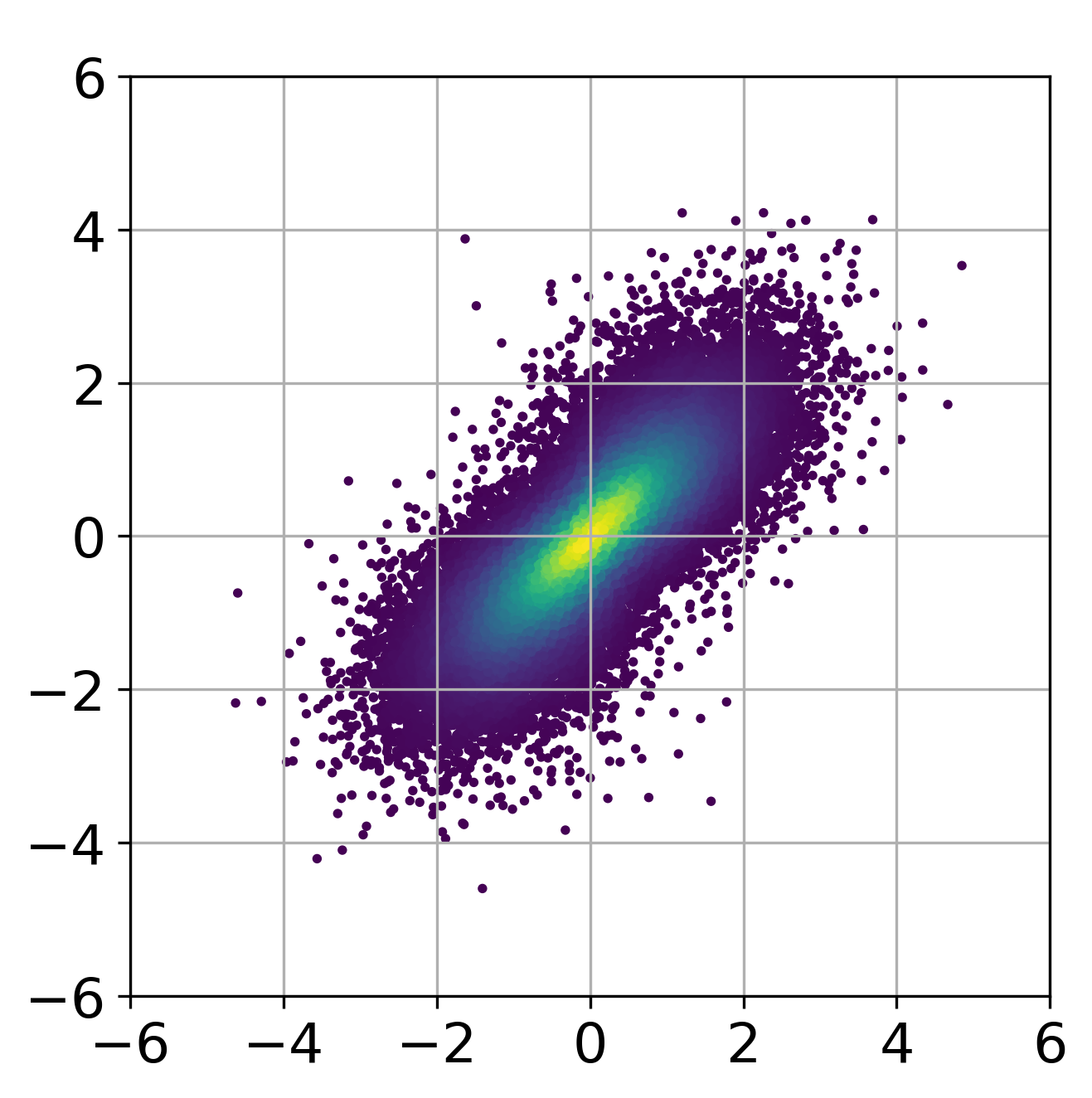}
            \caption{{\fontsize{8}{10}\selectfont cFM (SWD$=0.093$)}}
        \end{subfigure} &\hspace{-15pt}
        \begin{subfigure}{0.235\textwidth}
            \includegraphics[width=\linewidth]{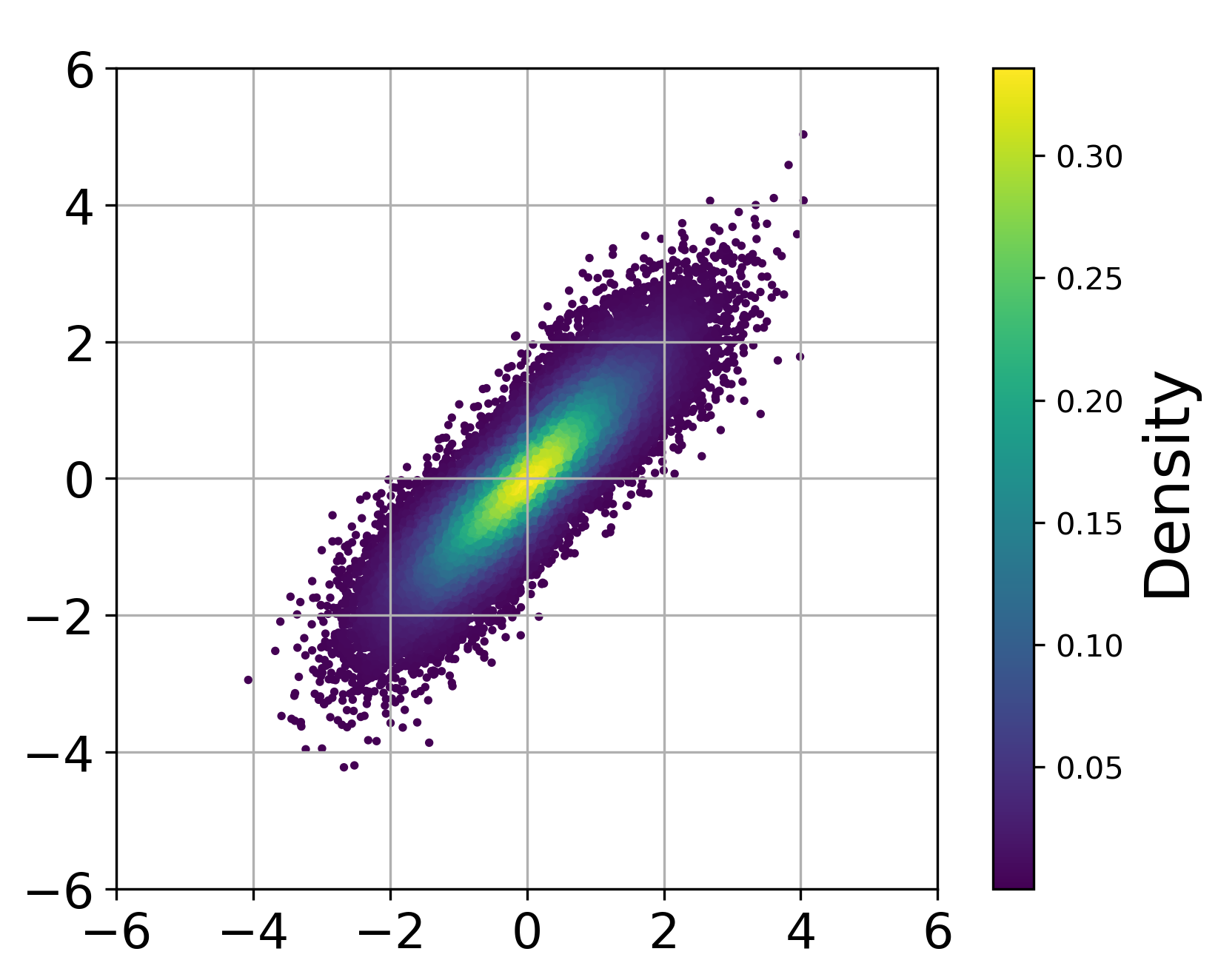}
            \caption{{\fontsize{8}{10}\selectfont cFM-$\gamma$ (SWD$=0.034$)}}
        \end{subfigure}
         \end{tabular}
\caption{Visualization of $40000$ samples from the prior ($\gamma = 0.9$) and the recovered distributions.}
    \label{fig:0p9}
\end{figure*}

 During the training stage, truth-observation pairs resulting from priors with $\gamma\in[-0.75,-0.25]\cup[0.25,0.75]$ are provided. At inference time, we evaluate the recovery performance for priors with $\gamma \in [-1,1]$ perturbed by \eqref{eq:toyforward}. Observation set size $N=4000$ and ensemble information dimension $k=3$ are set for EI-FM, i.e., $\phi_w:\mathbb{R}^{4000\times 2}\rightarrow \mathbb{R}^3$. Besides the mentioned baselines, we also evaluate cFM-$\gamma$, which is built based on cFM, but additionally conditioned on the latent information $\gamma$. To avoid overcrowding the results, we restrict the comparison to FM-based models. \begin{wrapfigure}{r}{0.5\textwidth}
        \centering
    \includegraphics[width =\linewidth]{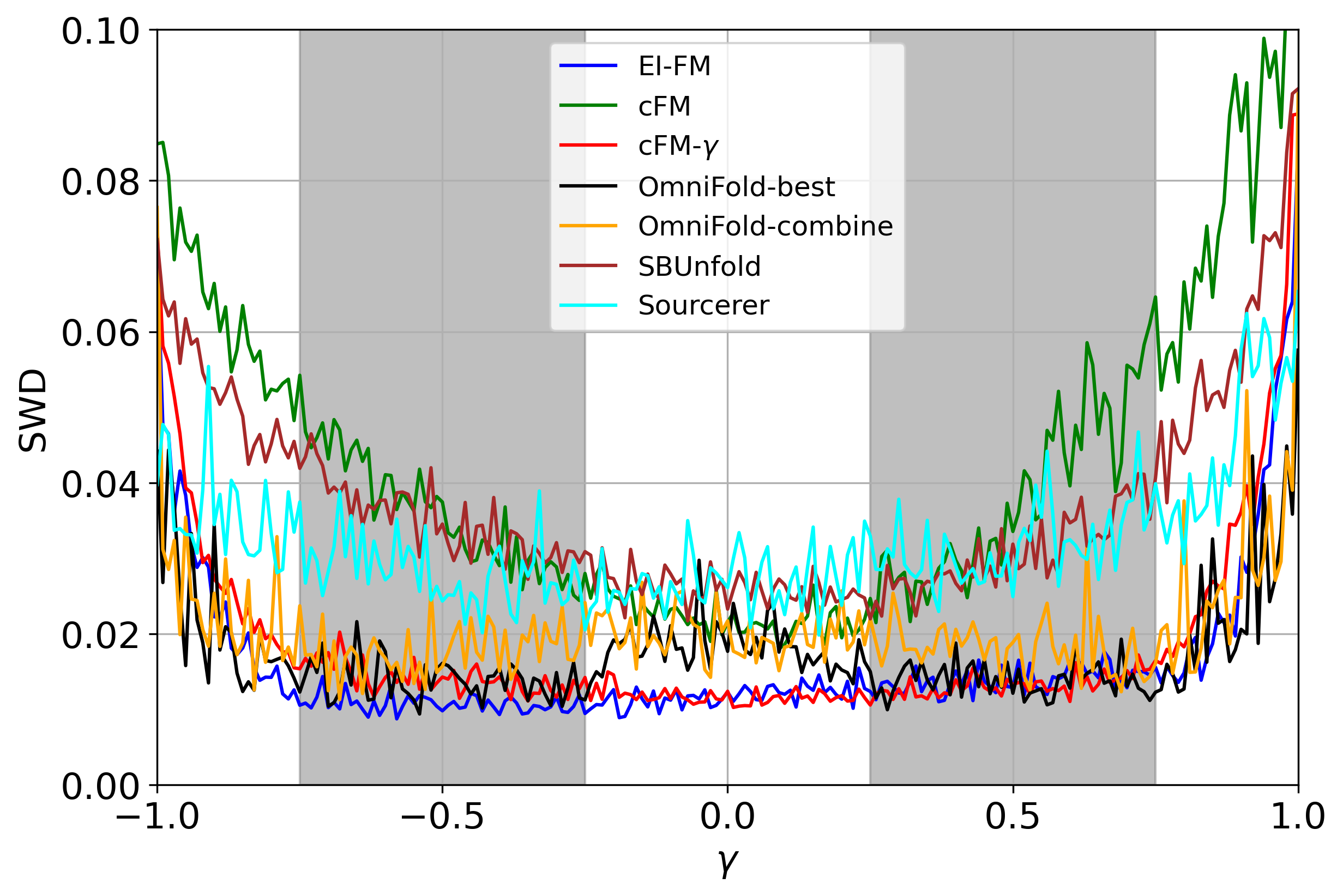}
    \caption{Average SWD($\downarrow$) between the truth and the recovery vs. $\gamma$, evaluated over $40000$ samples. Grey areas denote the priors contained in the training data.}
    \vspace{-20pt}
    \label{fig:2dres}
\end{wrapfigure}   GDDPM is excluded because it relies on physics-specific particle-moment information that is unavailable in this setting, while FM-$\gamma$ serves as the baseline using the optimal handcrafted summary statistic and assumes direct knowledge of the priors.

Fig.~\ref{fig:0p9} visualizes the distribution of the prior with $\gamma=0.9$ and the recovered distributions by $3$ representative methods to illustrate EI-FM's generalization ability for recovering priors in the same parameter family as in the training set.
The true prior with $\gamma=0.9$ is a ``thin'' distribution, which is unseen during training. cFM's recovery is much ``wider'' than the prior since it performs an element-wise generation without considering the ensemble information.
EI-FM, which incorporates the ensemble information from observed samples, can achieve similar performance to cFM-$\gamma$ with direct prior knowledge, illustrating its capability to generalize to unseen distributions.

In Fig.~\ref{fig:2dres}, we compare the SWD between the prior and the recovered distributions w.r.t. $40000$ samples for each $\gamma$ in $\{-1,-0.99,\cdots,0.99,1\}$ since SWD measures discrepancies between multivariate distributions efficiently.
EI-FM displays superior recovery performance among all compared methods and performs similarly to cFM-$\gamma$, for $\gamma\in[-1,1]$. OmniFold-best's initial distribution is exact the priors for $\gamma\in[-0.75,-0.25]\cup[0.25,0.75]$, leading to low SWD. However, OmniFold displays weaker generalization ability than EI-FM for $\gamma\in[-0.25,0.25]$.
Therefore, we can conclude that EI-FM is able to effectively utilize the ensemble information of observations to help infer the posterior and generalize to unseen distributions with performance comparable to models directly provided with prior information.

\begin{figure}[thb]
  \centering
  \includegraphics[width=0.8\textwidth]{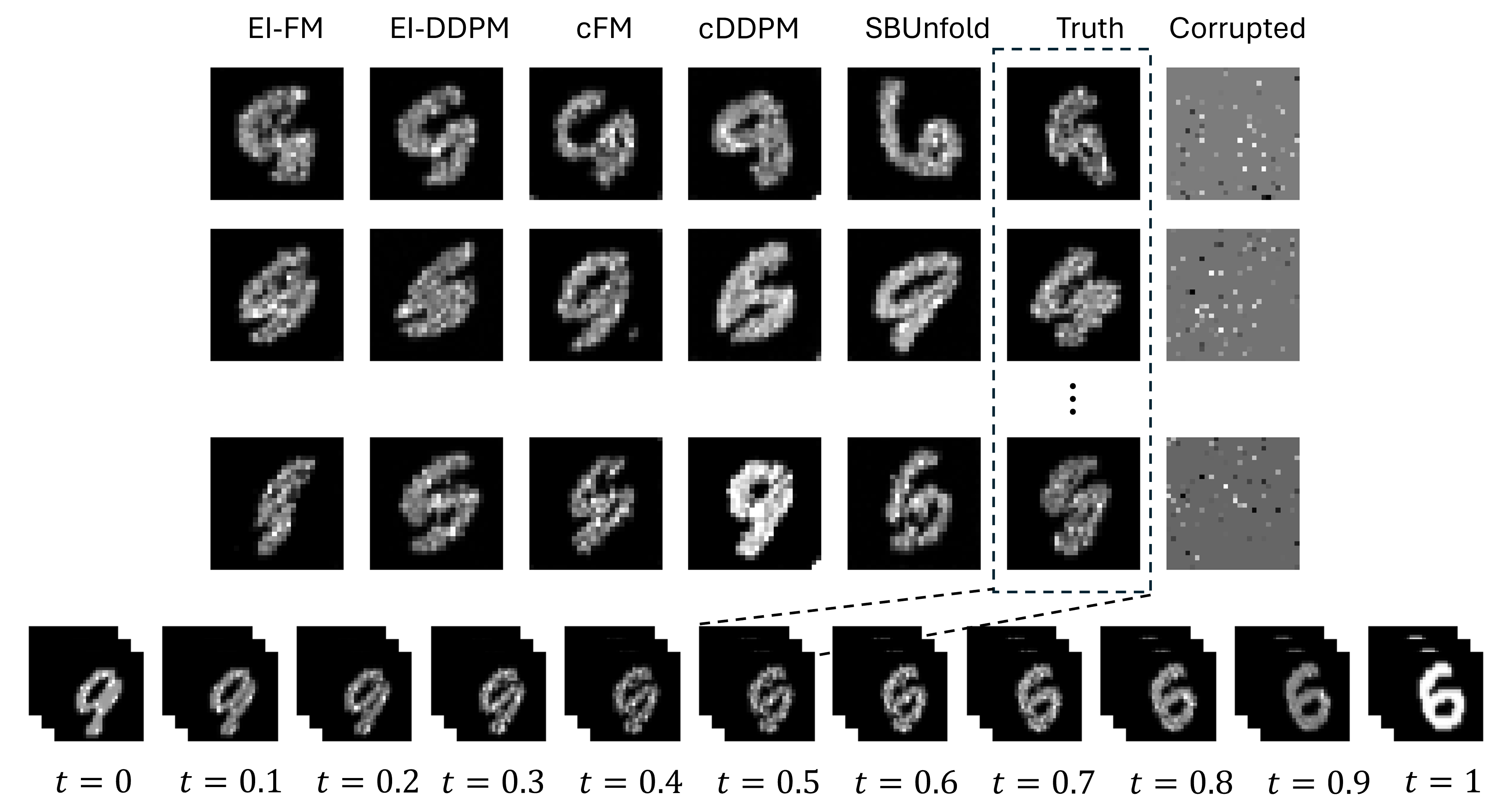}
  \caption{Upper: the recovered images via different methods, the truth ($t=0.5$), and the corrupted images. Lower: the transformation process from digit ``$9$'' to ``$6$''. At $t=0.5$, the ground-truth images exhibit a characteristic “S”-shaped structure, which EI-DDPM and EI-FM preserve more faithfully than the other methods.}
\label{fig:mixture_viz}
\end{figure}

    

\begin{figure}[htb]
    \centering
    \includegraphics[width=0.45\linewidth]{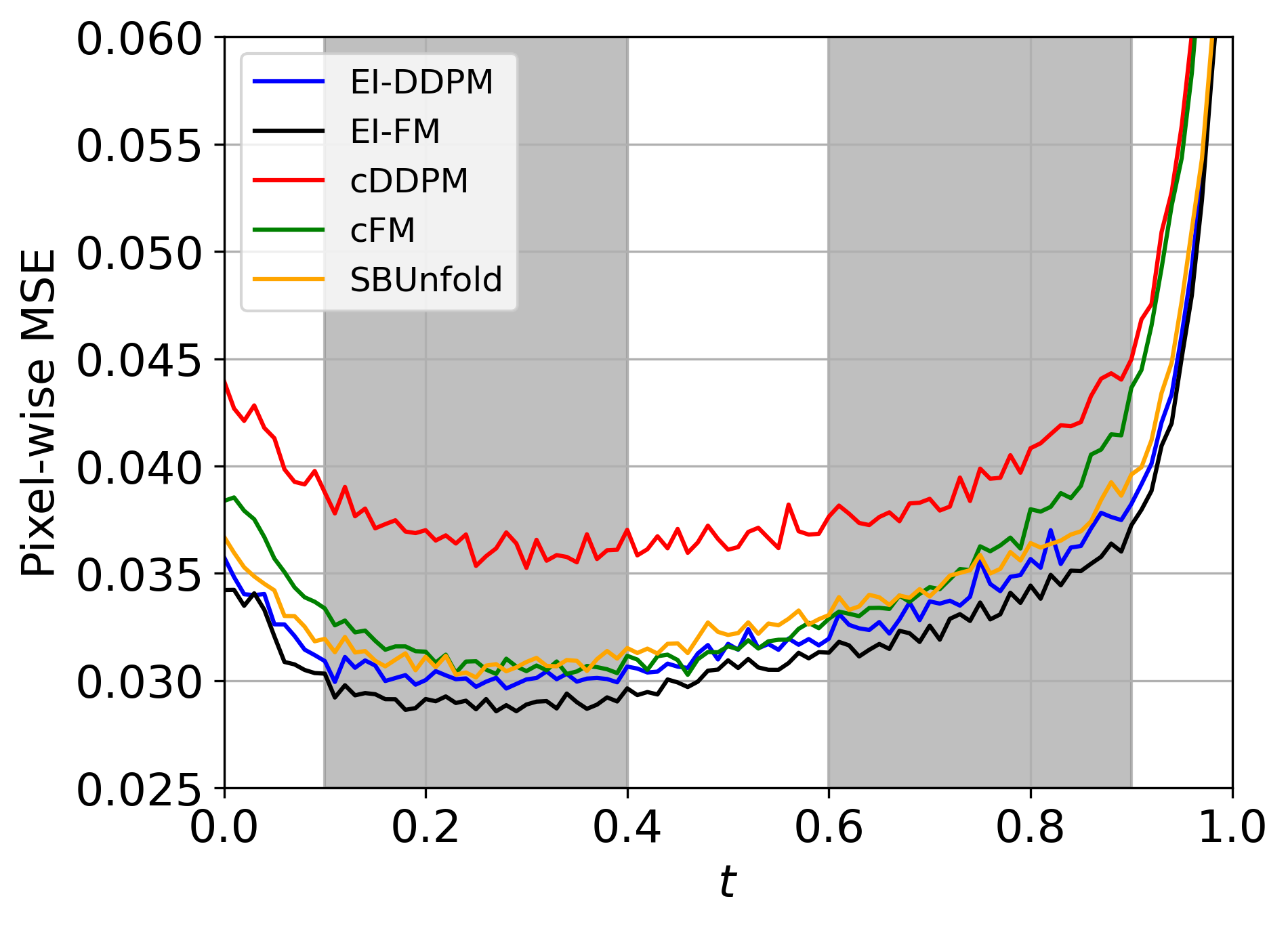}        \includegraphics[width=0.45\linewidth]{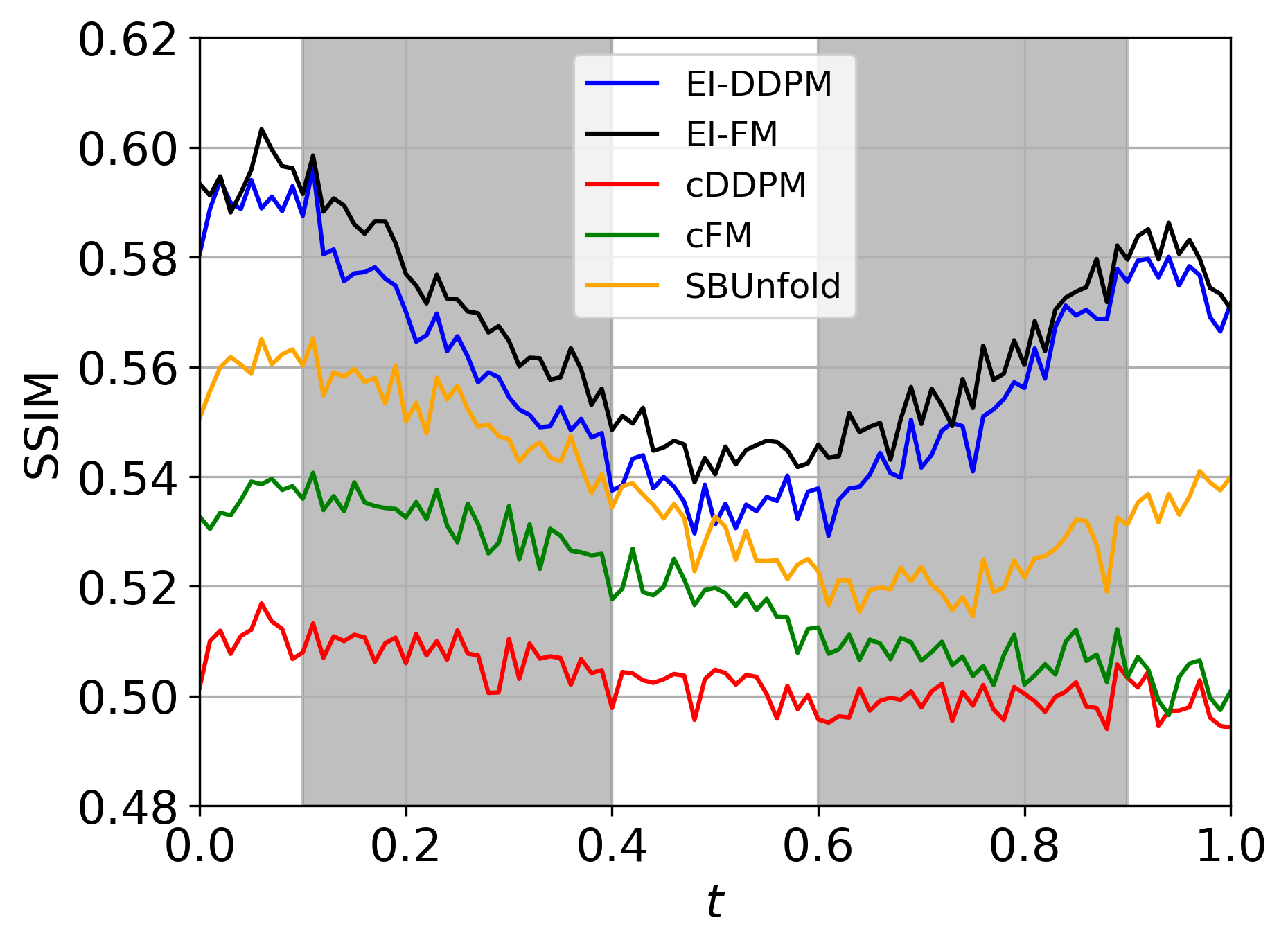}
    \caption{Pixel-wise MSE($\downarrow$) and SSIM($\uparrow$) vs. $t$. Grey areas denote the priors that are included in the training data. }
      \vspace{-10pt}
    \label{fig:mixture_stats}
\end{figure}

\subsection{Image Inversion of MNIST Digits Mixture}\label{sec:mnistexp}

As an example of a synthetic EIP on the image inversion task, we present an experiment on mixtures of MNIST digit images, where ensembles are generated by continuously interpolating between two digit classes and subsequently corrupted by stochastic pixel-wise noise. The goal is to recover the corresponding clean images for a set of corrupted observations with a common but unknown interpolation parameter. 

Specifically, the images of MNIST digit ``$9$'' continuously transform into MNIST digit ``$6$'' over time $t\in[0,1]$. The images are all ``$9$'' at $t=0$ and become ``$6$'' at $t=1$. For $0<t<1$, the images are mixtures of the two digits, resembling ``$6$'' more and ``$9$'' less as $t$ approaches $1$. 
The process of creating the mixture of two MNIST digits following \citet{WFM} is described as follows. First, the MNIST digit images are converted to point clouds. Then an entropically regularized optimal transport (OT) plan between two weighted point clouds is computed using OTT’s Sinkhorn solver, producing a soft matching matrix. 
Based on the matrix, greedy “rounded matching” is applied by repeatedly selecting the maximum probability entry in the matrix, assigning that source to the corresponding target, and zeroing out the associated row and column to prevent reuse. This process iterates until all points are matched, leading to a permutation-like hard assignment that approximates the true optimal permutation matrix implied by the OT solution. The resulting hard assignment defines a transport path parameterized by time $t$, where $t=0$ corresponds to the initial point clouds and $t=1$ corresponds to the target point clouds. The intermediate $t$ 
interpolates each point along its assigned displacement toward its target. Finally, the point clouds are converted back to images. At the inference stage, given a set of corrupted images, which come from a common prior at an unknown interpolation time $t$, the objective is to recover the corresponding clean images for each corrupted image in the set.

Let $x_{a,b}$ denote the $a$-th rows' $b$-th pixel value in a MNIST image $x\in\mathbb{R}^{28\times 28}$. We apply an element-wise corruption designed to make recovery from a single observation difficult while preserving prior information at the observation set level, and the forward process is given by: 
\begin{equation}
\begin{aligned}
    &y_{a,b} = x_{a,b}+n(x_{a,b}),
    \quad n(x_{a,b})\sim\begin{cases}
        \delta_{-x_{a,b}},&\text{with probability }0.9;
        \\\mathcal{N}(0,2), &\text{with probability } 0.1.
    \end{cases}
    \end{aligned}
\end{equation}

Setting $N=128$ and $\phi_w:\mathbb{R}^{128\times28\times28}\rightarrow \mathbb{R}^{28\times 28}$ for EI-DDPM and EI-FM, we compare our proposed methods with cFM, cDDPM, and SBUnfold for the image inversion task. 
Each method is provided with pairs of clean images and corrupted images resulting from priors with $t\in[0.1,0.4]\cup[0.6,0.9]$. At inference time, each method aims to recover the original images from a set of images with the same but unknown $t$.

This image inverse problem requires posterior recovery; therefore, prior-only methods such as OmniFold and Sourcerer are not applicable. Similar to the case in  Sec.~\ref{sec:gaussian}, GDDPM is again excluded for comparison, because the required physics-specific particle-moment information is unavailable in this setting.

First, we visualize the recovery performance for $t=0.5$ in Fig.~\ref{fig:mixture_viz}. EI-FM and EI-DDPM preserve the “S”-shaped structural patterns of the truth much more faithfully, whereas the other baselines exhibit clear structural distortions and fail to recover the correct geometry. Then we sweep $t\in[0,1]$ with an interval $0.01$ and use pixel-wise mean squared error (MSE) to measure pixel-wise accuracy and the structural similarity index measure (SSIM) to assess structural fidelity. The results are shown in Fig.~\ref{fig:mixture_stats}. Both regions near $t=0$ and $t=1$ are excluded from training. Numerical errors in the asymmetrically constructed transport path accumulate from $t=0$ toward $t=1$. Samples near $t=1$ therefore deviate more from the original digit distribution, making generalization more difficult and increasing the MSE.
Fig.~\ref{fig:mixture_stats} demonstrates EI-FM and EI-DDPM's superior performance in both MSE and SSIM, indicating that EI-FM and EI-DDPM can scale up to high-dimensional settings and effectively incorporate the ensemble information for posterior inferences and generalizations. 


\begin{figure*}[htb]
    \centering
    \includegraphics[width=0.31\linewidth]{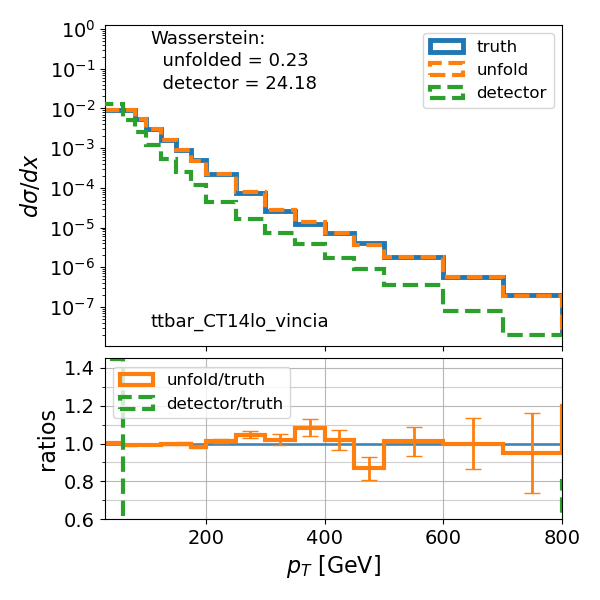}
    \includegraphics[width=0.31\linewidth]{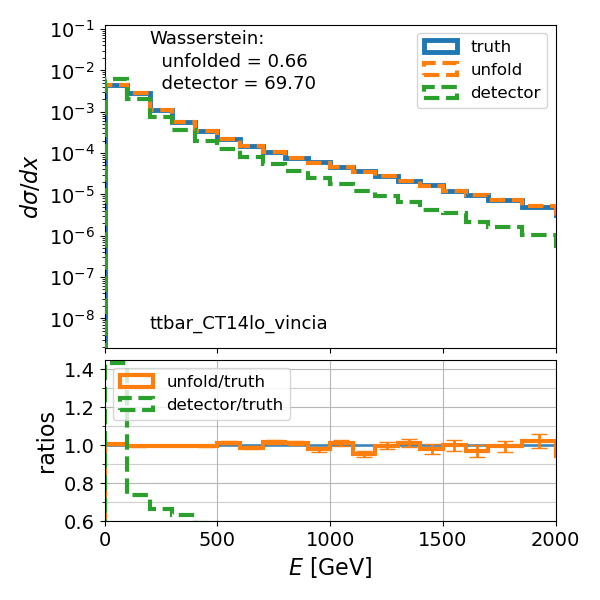}
    \includegraphics[width=0.31\linewidth]{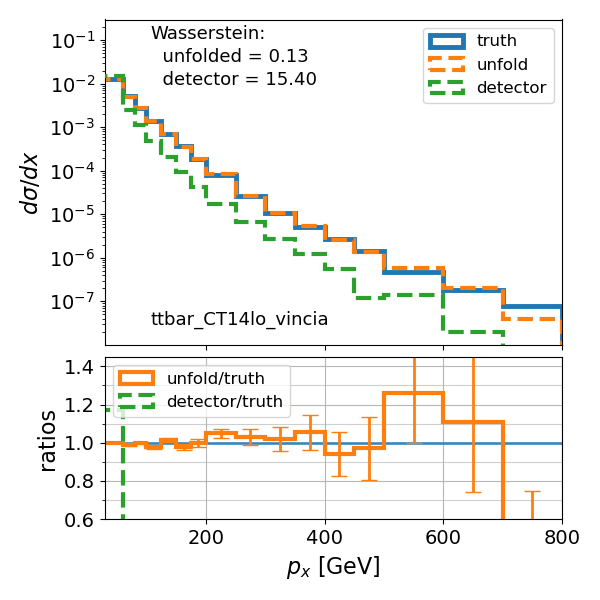}
    \caption{ Unfolding results of jet kinematics from a $t\bar{t}$ process (modeled with the CT14lo PDF and Vincia parton showers) from the data-driven detector smearing using EI-FM. }
    \vspace{-5pt}
    \label{fig:physicsttbar}
\end{figure*}
\begin{table*}[htb]
\centering
\resizebox{\textwidth}{!}{
\begin{tabular}{|c|c|c|c|c|c|c|c|c|c|c|c|c|}
\hline
WD ($\downarrow$) & Name & Detector & EI-DDPM & EI-FM & cDDPM & cFM & GDDPM-v & GDDPM & \begin{tabular}[c]{@{}c@{}}OmniFold\\ -best\end{tabular} & \begin{tabular}[c]{@{}c@{}}OmniFold\\ -combine\end{tabular} & SBUnfold &   Sourcerer \\ \hline
\multirow{4}{*}{$p_T$} & Leptoquark & $31.85$ & $0.44$ & $0.44$ & $1.08$ & $2.65$ & $0.73$ & $0.44$ & $\textcolor{red}{\mathbf{0.19}}$ & $0.82$ & $18.10$&$7.96$  \\ \cline{2-13} 
 & $t\bar{t}$ (CT14lo, Vincia) & $24.18$ & $0.44$ & $\textcolor{red}{\mathbf{0.23}}$ & $1.01$ & $3.36$ & $1.36$ & $0.55$ & $0.60$ & $0.52$ & $1.88$ &$15.14$ \\ \cline{2-13} 
 & $W+$jets (CT14lo) & $18.60$ & $0.60$ & $0.44$ & $2.41$ & $4.14$ & $0.53$ & $0.48$ & $\textcolor{red}{\mathbf{0.11}}$ & $0.37$ & $21.07$&$25.96$  \\ \cline{2-13} 
 & $Z+$jets (CTEQ6L1) & $15.81$ & $0.51$ & $\textcolor{red}{\mathbf{0.45}}$ & $2.55$ & $5.55$ & $2.25$ & $1.98$ & $0.48$ & $0.64$ & $25.18$ &$19.59$ \\ \hline
\multirow{4}{*}{$E$} & Leptoquark & $83.87$ & $\textcolor{red}{\mathbf{0.46}}$ & $0.76$ & $4.70$ & $2.66$ & $1.47$ & $0.63$ & $1.08$ & $0.47$ & $13.99$ &$57.42$  \\ \cline{2-13} 
 & $t\bar{t}$ (CT14lo, Vincia) & $69.70$ & $0.77$ & $\textcolor{red}{\mathbf{0.66}}$ & $2.96$ & $3.29$ & $1.54$ & $0.89$ & $0.83$ & $1.41$ & $4.83$ &$104.13$  \\ \cline{2-13} 
 & $W+$jets (CT14lo) & $90.42$ & $1.08$ & $1.60$ & $4.56$ & $4.64$ & $3.38$ & $1.60$ & $\textcolor{red}{\mathbf{0.56}}$ & $3.05$ & $23.67$ &$94.95$  \\ \cline{2-13} 
 & $Z+$jets (CTEQ6L1) & $83.18$ & $\textcolor{red}{\mathbf{0.81}}$ & $1.19$ & $6.83$ & $12.62$ & $6.25$ & $7.04$ & $1.44$ & $2.22$ & $40.67$ &$79.37$ \\ \hline
\multirow{4}{*}{$p_x$} & Leptoquark & $20.26$ & $\textcolor{red}{\mathbf{0.21}}$ & $0.26$ & $0.95$ & $1.39$ & $0.73$ & $0.25$ & $0.41$ & $0.43$ & $10.53$ &$5.89$  \\ \cline{2-13} 
 & $t\bar{t}$ (CT14lo, Vincia) & $15.40$ & $0.19$ & $\textcolor{red}{\mathbf{0.13}}$ & $0.65$ & $1.03$ & $0.82$ & $0.31$ & $0.41$ & $0.30$ & $1.00$ &$11.22$  \\ \cline{2-13} 
 & $W+$jets (CT14lo) & $11.84$ & $0.26$ & $\textcolor{red}{\mathbf{0.21}}$ & $1.07$ & $0.90$ & $0.75$ & $\textcolor{red}{\mathbf{0.21}}$ & $0.24$ & $0.22$ & $8.75$ &$11.04$  \\ \cline{2-13} 
 & $Z+$jets (CTEQ6L1) & $10.06$ & $0.23$ & $\textcolor{red}{\mathbf{0.19}}$ & $1.19$ & $0.66$ & $1.22$ & $1.07$ & $0.35$ & $0.31$ & $16.08$ &$10.74$ \\ \hline
\end{tabular}}
\caption{Result of data recovery performance on $4$ unseen physics distributions. We report
    the $1$-D Wasserstein distance between the truth-level data and detector-level data / recovered data via various methods for $p_T,E$ and $p_x$ (complete results in Sec.~\ref{sec:unfoldingfull}). The best results are noted in red. }
    \vspace{-5pt}
    \label{tb:physicsres}
\end{table*}
\subsection{Particle Physics Data Unfolding}\label{sec:unfold}
In this section, we evaluate the proposed methods on simulated particle physics data. The data consists of quantum chromodynamics (QCD) jets, which are collimated sprays of particles produced when partons (the constituent particles within protons) fragment in high-energy collisions. These datasets are generated using the PYTHIA 8.3 event generator \cite{pythiaManual} for various physics processes such as $t\bar{t}$, $W+$jets, $Z+$jets, dijet, and leptoquark processes. The jet kinematics include transverse momentum ($p_T$), pseudorapidity ($\eta$),  azimuthal angle ($\phi$), and $4$-momentum components ($E,p_x,p_y,p_z$).
These jets are presented at $2$ stages: the truth-level ($x$) representation is constructed from the direct output of the Monte Carlo event generator, while the detector-level ($y$) is the representation after the jets pass through the detector simulation.
The training data consists of paired truth-level and detector-level jet vectors. The truth-level jet vectors come from $18$ different physics processes, including various parton distribution functions and parton shower models, and the detector effects are identical across all truth-level jet vectors.  We refer readers to \citet{Camila24} for more details on this dataset.
During inference time, we compare the distribution similarity between the recovered data from $4$ unseen physics processes and their truth-level data.

GDDPM \citep{Camila24} proposes to incorporate the first six moments of the $p_T$ to help with unfolding. However, this implicitly assumes that $p_T$ contains the complete distributional information of the $7$-component vector. Therefore, we also consider a more general variant, referred to as GDDPM-v, in which this assumption is not made and moments of all $7$ components are taken as the conditional information. The Wasserstein-$1$ distance (WD) \citep{WD1} for each jet kinematics between the true distributions and the recovered distributions is selected as the metric for measuring the distribution similarity following \citet{Camila24}.

$N=2000$ and $\phi_w:\mathbb{R}^{2000\times 7}\rightarrow \mathbb{R}^{6}$ are fixed in both EI-DDPM and EI-FM in this particle physics unfolding task. Fig.~\ref{fig:physicsttbar} showcases the EI-FM's reconstruction of $p_T,E$ and $p_x$ distributions from a $t\bar{t}$ process. The detector effects cause a great difference between the truth and the detector-level distributions. EI-FM is able to recover distributions with small WDs to the truth.
Table~\ref{tb:physicsres} shows the recovery performance of $p_T$, $E$ and $p_x$ for $4$ unseen physics processes.
The proposed methods display superior performance across all $4$ unseen physics processes, illustrating the effectiveness of the proposed methods in utilizing latent ensemble information for unfolding without knowledge of the priors. It is worth mentioning that GDDPM outperforms GDDPM-v, suggesting that redundant moment information in GDDPM-v impairs recovery. Nevertheless, our proposed methods achieve comparable or superior performance to GDDPM, indicating that $\phi_w(\cdot)$ can automatically extract the core ensemble information from $\mathcal{Y}$ and eliminate redundant information. Posterior accuracy is also evaluated in Sec.~\ref{app:tarp} using the Test of Accuracy with Random Points (TARP) metric for both the $2$-D Gaussian EIP and the unfolding task.



\begin{table}[thb]
\centering
\scalebox{0.7}{
\begin{tabular}{|c|c|c|c|c|c|c|c|}
\hline
 Families & Metric & EI-DDPM & EI-FM & cDDPM & cFM & InvNet & VGAN \\
\hline
\multirow{3}{*}{FlatVel-A} & MSE($\downarrow$) & $\textcolor{red}{\mathbf{0.012}}$ & $\textcolor{red}{\mathbf{0.012}}$ & $0.020$ & $0.013$ & $0.017$ & $0.018$ \\
 & MAE($\downarrow$) & $\textcolor{red}{\mathbf{0.042}}$ & $0.045$ & $0.062$ & $0.049$ & $0.068$ & $0.069$ \\
 & SSIM($\uparrow$) & $\textcolor{red}{\mathbf{0.908}}$ & $0.905$ & $0.869$ & $0.898$ & $0.879$ & $0.881$ \\
\hline
\multirow{3}{*}{FlatVel-B} & MSE($\downarrow$) & $\textcolor{red}{\mathbf{0.119}}$ & $0.130$ & $0.129$ & $0.139$ & $0.154$ & $0.158$ \\
 & MAE($\downarrow$) & $\textcolor{red}{\mathbf{0.139}}$ & $0.151$ & $0.160$ & $0.154$ & $0.196$ & $0.199$ \\
 & SSIM($\uparrow$) & $\textcolor{red}{\mathbf{0.774}}$ & $0.762$ & $0.759$ & $0.753$ & $0.701$ & $0.694$ \\
\hline
\multirow{3}{*}{CurveVel-A} & MSE($\downarrow$) & $0.118$ & $\textcolor{red}{\mathbf{0.112}}$ & $0.117$ & $0.115$ & $0.131$ & $0.139$ \\
 & MAE($\downarrow$) & $0.214$ & $\textcolor{red}{\mathbf{0.202}}$ & $0.215$ & $0.213$ & $0.221$ & $0.229$ \\
 & SSIM($\uparrow$) & $0.564$ & $\textcolor{red}{\mathbf{0.584}}$ & $0.545$ & $0.551$ & $0.555$ & $0.544$ \\
\hline
\multirow{3}{*}{CurveVel-B} & MSE($\downarrow$) & $\textcolor{red}{\mathbf{0.334}}$ & $0.345$ & $0.427$ & $0.347$ & $0.445$ & $0.454$ \\
 & MAE($\downarrow$) & $\textcolor{red}{\mathbf{0.307}}$ & $0.319$ & $0.366$ & $0.324$ & $0.394$ & $0.393$ \\
 & SSIM($\uparrow$) & $\textcolor{red}{\mathbf{0.526}}$ & $0.504$ & $0.472$ & $0.497$ & $0.419$ & $0.422$ \\
\hline
\multirow{3}{*}{FlatFault-A} & MSE($\downarrow$) & $\textcolor{red}{\mathbf{0.013}}$ & $0.014$ & $0.023$ & $0.015$ & $0.023$ & $0.026$ \\
 & MAE($\downarrow$) & $\textcolor{red}{\mathbf{0.032}}$ & $0.036$ & $0.054$ & $0.041$ & $0.061$ & $0.071$ \\
 & SSIM($\uparrow$) & $\textcolor{red}{\mathbf{0.966}}$ & $0.962$ & $0.958$ & $0.961$ & $0.945$ & $0.942$ \\
\hline
\multirow{3}{*}{FlatFault-B} & MSE($\downarrow$) & $0.171$ & $0.171$ & $0.186$ & $\textcolor{red}{\mathbf{0.169}}$ & $0.184$ & $0.184$ \\
 & MAE($\downarrow$) & $\textcolor{red}{\mathbf{0.268}}$ & $0.275$ & $0.285$ & $0.273$ & $0.288$ & $0.287$ \\
 & SSIM($\uparrow$) & $\textcolor{red}{\mathbf{0.546}}$ & $0.531$ & $0.527$ & $0.531$ & $0.504$ & $0.497$ \\
\hline
\multirow{3}{*}{CurveFault-A} & MSE($\downarrow$) & $\textcolor{red}{\mathbf{0.026}}$ & $0.029$ & $0.035$ & $0.030$ & $0.045$ & $0.055$ \\
 & MAE($\downarrow$) & $\textcolor{red}{\mathbf{0.054}}$ & $0.062$ & $0.075$ & $0.065$ & $0.095$ & $0.113$ \\
 & SSIM($\uparrow$) & $\textcolor{red}{\mathbf{0.939}}$ & $0.932$ & $0.923$ & $0.928$ & $0.904$ & $0.901$ \\
\hline
\multirow{3}{*}{CurveFault-B} & MSE($\downarrow$) & $0.353$ & $\textcolor{red}{\mathbf{0.351}}$ & $0.360$ & $0.363$ & $0.383$ & $0.407$ \\
 & MAE($\downarrow$) & $\textcolor{red}{\mathbf{0.413}}$ & $0.415$ & $0.420$ & $0.426$ & $0.432$ & $0.442$ \\
 & SSIM($\uparrow$) & $\textcolor{red}{\mathbf{0.346}}$ & $0.341$ & $0.332$ & $0.338$ & $0.306$ & $0.300$ \\
\hline
\multirow{3}{*}{Style-A} & MSE($\downarrow$) & $\textcolor{red}{\mathbf{0.090}}$ & $0.097$ & $0.092$ & $0.095$ & $0.093$ & $0.097$ \\
 & MAE($\downarrow$) & $\textcolor{red}{\mathbf{0.198}}$ & $0.210$ & $0.201$ & $0.208$ & $0.210$ & $0.213$ \\
 & SSIM($\uparrow$) & $0.583$ & $0.554$ & $\textcolor{red}{\mathbf{0.586}}$ & $0.563$ & $0.557$ & $0.557$ \\
\hline
\multirow{3}{*}{Style-B} & MSE($\downarrow$) & $\textcolor{red}{\mathbf{0.047}}$ & $0.050$ & $0.048$ & $0.050$ & $0.051$ & $0.051$ \\
 & MAE($\downarrow$) & $\textcolor{red}{\mathbf{0.157}}$ & $0.163$ & $0.157$ & $0.161$ & $0.162$ & $0.163$ \\
 & SSIM($\uparrow$) & $\textcolor{red}{\mathbf{0.559}}$ & $0.541$ & $\textcolor{red}{\mathbf{0.559}}$ & $0.547$ & $0.544$ & $0.546$ \\
\hline
\end{tabular}
}
\caption{MSE, MAE, and SSIM of the compared methods in the FWI task. InvNet and VGAN denote InversionNet and VelocityGAN, respectively. The best results are denoted in red.}
\label{table:FWIres}
\vspace{-10pt}
\end{table}

\subsection{Seismic Full Waveform Inversion}

In this section, we study the $2$D full waveform inversion (FWI) task \citep{NEURIPS2022_FWI}, in which the goal is to reconstruct subsurface velocity maps of various families from seismic waveform recordings. Seismic data is the observation of wave propagation governed by the acoustic wave equation, while the corresponding velocity models serve as inversion targets. The FWI datasets are organized in several families, each representing a distinct class of subsurface structures. These families can be naturally interpreted as structural priors in the inverse problem. In our experiments, models are trained on $8$ families: FlatVel-A, FlatVel-B, CurveVel-B, FlatFault-A, FlatFault-B, CurveFault-A, Style-A, and Style-B. Performance is evaluated on unseen samples from these training families as well as on $2$ entirely unseen families, CurveVel-A and CurveFault-B, in order to assess the models’ generalization capability across both data instances and structural priors. At the inference stage, given a set of seismic data from a common family, the objective is to recover each subsurface velocity map in the set.

We downsample both the seismic data and velocity models along the temporal and spatial dimensions and each input sample is a tensor of shape $(5,128,32)$, corresponding to seismic recordings from $5$ source locations, $128$ temporal samples, and $32$ receiver positions along the surface, while the output is a velocity map of shape $(1,32,32)$, representing a single-channel subsurface velocity map discretized on a $32\times 32$ spatial grid.

$N=64$ and $\phi_w:\mathbb{R}^{64\times5\times128\times32}\rightarrow\mathbb{R}^{32\times 32}$ are fixed for our proposed methods in this task. As this FWI task requires posterior recovery, prior-targeting methods are not applicable. We therefore compare against two representative FWI-specific methods, InversionNet \citep{WuYue2020IAEa} and VelocityGAN \citep{velocitygan}, as well as the general conditional generative baselines cDDPM and cFM. Visualizations of the recovered velocity maps from each family via various methods are shown in Appendix~\ref{sec:visFWI}.
Following standard FWI evaluation protocols, we use MSE and MAE to quantify point-wise velocity errors and SSIM to assess recovered spatial structures. As shown in Table~\ref{table:FWIres}, EI-DDPM and EI-FM achieve the best results in most family--metric comparisons and demonstrate strong generalization on the CurveVel-A and CurveFault-B families, which are excluded from the training priors. These results indicate that the proposed methods scale effectively to practical high-dimensional recovery tasks and exhibit robust generalization capability with the ensemble information.

\section{Conclusions and Limitations}

We introduce EIP, in which one aims to invert for an ensemble distributed according to the pushforward of an unknown prior under a forward process. To address this problem, we propose ensemble inverse generative models, a posterior sampling framework that conditions on both the individual observation and ensemble information extracted from the observation set by a permutation-invariant neural network. Using information from the observation ensemble, the proposed method turns a standard sample-wise posterior sampler into a distribution-dependent posterior sampler. This allows the model to exploit information about the unknown prior that is implicitly contained in the observation set, providing an inductive bias for generalization to unseen priors. Empirically, EI-DDPM and EI-FM demonstrate strong posterior inference and generalization performance across inverse imaging, HEP unfolding, and FWI. The framework also has limitations: its performance depends on the informativeness of the observation set and on the quality, capacity, and stability of the permutation-invariant encoder, whose optimal design is not addressed in this work. Future directions include further validation on directly measured data in real-world inverse problems, theoretical guarantees on the recovery discrepancy, and improved architectures for ensemble information extraction.

\newpage

\section*{Acknowledgments}

This work has been made possible thanks to the support of the Department of Energy Office of Science through the Grant DE-SC0023964. Zhengyan Huan and Shuchin Aeron are supported via NSF under Cooperative Agreement PHY-2019786. Shuchin Aeron would also like to acknowledge funding via NSF DMS 2309519.

\bibliography{tmlr}
\bibliographystyle{tmlr}

\newpage
\appendix
\onecolumn

\section{Additional Illustration and Notation}
\label{app:illustration_notation}

\subsection{Relationship between EIP-I and EIP-II}
\label{app:eip_relationship}
\begin{figure*}[htb]
    \centering
    
    \begin{subfigure}[b]{0.31\textwidth}
        \centering
        \includegraphics[width=\textwidth]{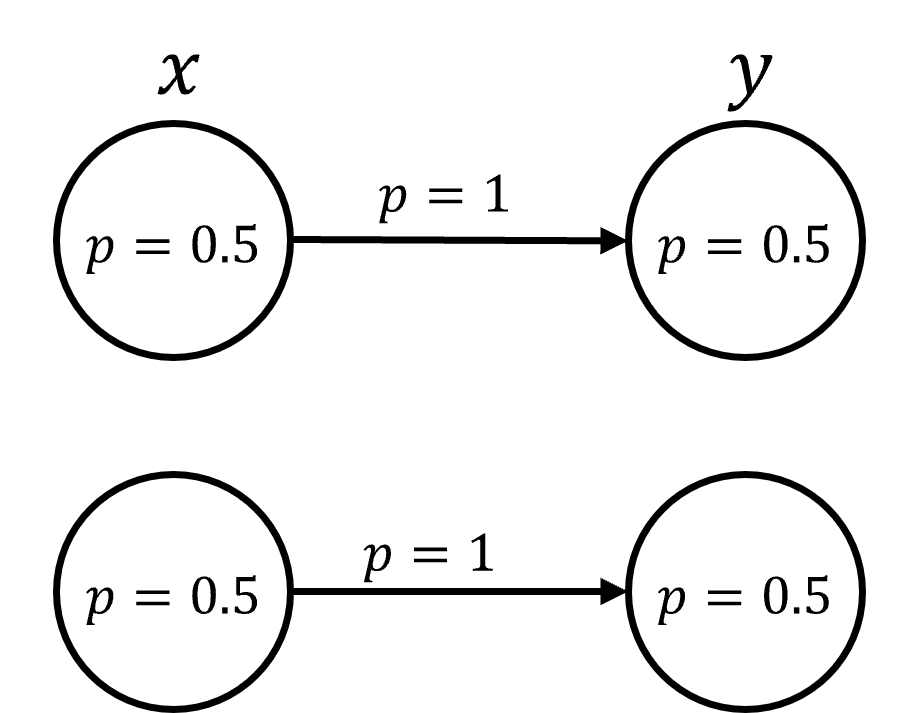}
        \caption{Forward process}
        \label{fig:sub1}
    \end{subfigure}
    \begin{subfigure}[b]{0.31\textwidth}
        \centering
        \includegraphics[width=\textwidth]{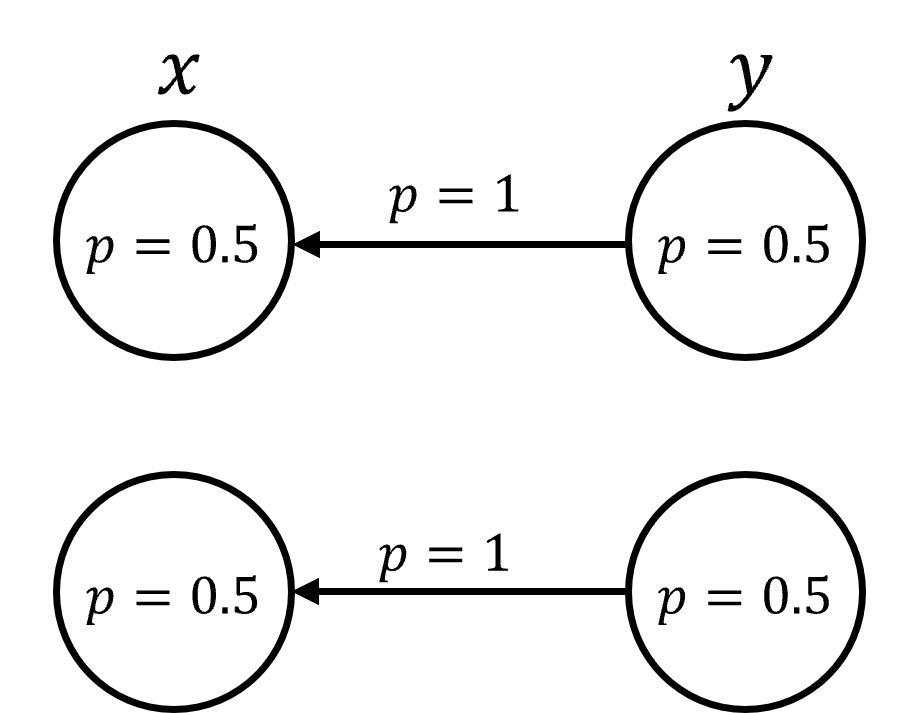}
        \caption{EIP-II's solution}
        \label{fig:sub2}
    \end{subfigure}
        \begin{subfigure}[b]{0.31\textwidth}
        \centering
        \includegraphics[width=\textwidth]{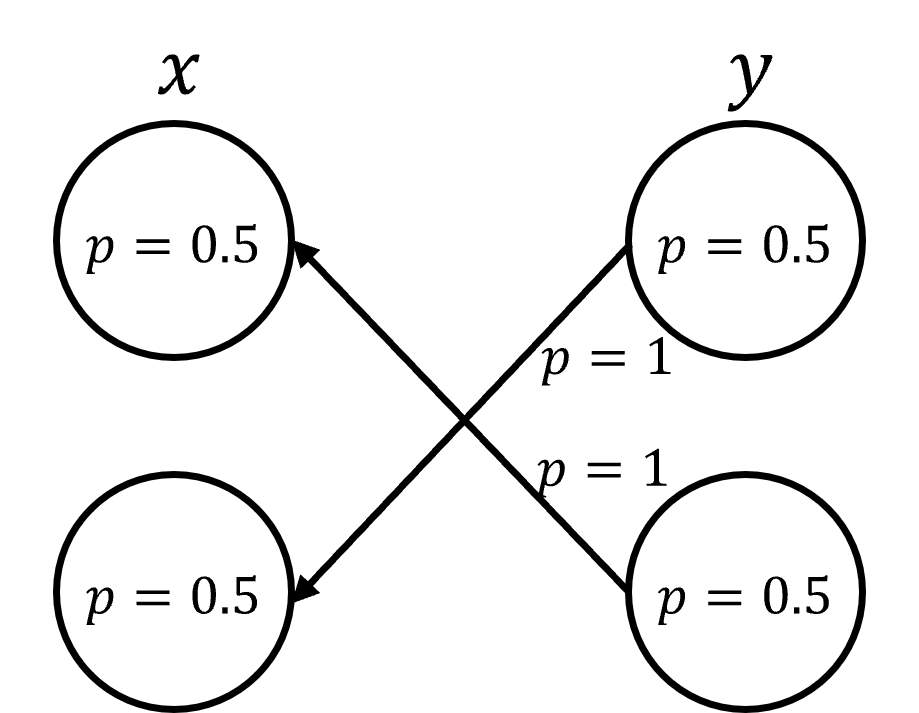}
        \caption{An incorrect posterior }
        \label{fig:sub3}
    \end{subfigure}
    \caption{Consider a forward process in Fig.~\ref{fig:sub1}, Fig.~\ref{fig:sub2} shows EIP-II's solution, with its integration corresponding to EIP-I's solution. Fig.~\ref{fig:sub3} shows an incorrect posterior; however, the integration of this incorrect posterior can lead to the correct prior. }
    \label{fig:introf}
\end{figure*}

\subsection{Summary of Notation}
\vspace{-10pt}
\begin{longtable}{
    >{\centering\arraybackslash}p{0.18\textwidth}
    p{0.76\textwidth}
}

\label{tab:notation}\\
\toprule
\textbf{Symbol} & \textbf{Explanation} \\
\midrule
\endfirsthead

\toprule
\textbf{Symbol} & \textbf{Explanation} \\
\midrule
\endhead

\endfoot


$\x$ &
Unknown truth variable to be recovered. \\

$\y$ &
Individual observation associated with $\x$ through the forward model. \\

$d$ &
Dimension of the variable under consideration. \\

$F$ &
Forward or measurement operator. \\

$n(\x)$ &
Additive noise, which may depend on $\x$. \\

$M$ &
Number of training datasets or training priors. \\

$\mathcal{D}$ &
Collection of paired training datasets,
$\mathcal{D}=\{\mathcal{D}_1,\ldots,\mathcal{D}_M\}$. \\

$(\x^{m,j},\y^{m,j})$ &
The $j$-th truth--observation pair in $\mathcal{D}_m$. \\

$\mathcal{D}_m$ &
Training dataset generated from the $m$-th prior,
$\mathcal{D}_m=\{(\x^{m,j},\y^{m,j})\}_{j=1}^{N_m}$. \\

$p(x)$ &
Unknown target prior distribution. \\

$p_m(x)$ &
Prior distribution associated with the $m$-th training dataset. \\

$\mathcal{Y}$ &
Observation ensemble from an unknown target prior,
$\mathcal{Y}=\{y^1,\ldots,y^N\}$. \\

$N$ &
Number of observations used to construct the ensemble information. \\

$N'$ &
Number of observations available at inference time when it differs from $N$. \\

$\phi_w$ &
Permutation-invariant neural network that extracts ensemble information from $\mathcal{Y}$. \\

$k$ &
Dimension of the learned ensemble information $\phi_w(\mathcal{Y})$. \\

$\varepsilon_\theta$ &
Neural network that predicts the noise in EI-DDPM or the velocity field in EI-FM. \\

$\x_t$ &
Intermediate state of the diffusion or flow process at time $t$. \\

$t$ &
Diffusion step in EI-DDPM or continuous flow time in EI-FM. \\

$\eta$ &
Learning rate used during training. \\

$\widehat{x}^{\,j}$ &
Recovered truth sample corresponding to observation $y^j$. \\

$T$ &
Total number of diffusion steps in EI-DDPM. \\

$\beta_t$ &
Forward-process noise variance at diffusion step $t$. \\

$\alpha_t$ &
Diffusion coefficient defined by $\alpha_t=1-\beta_t$. \\

$\bar{\alpha}_t$ &
Cumulative diffusion coefficient,
$\bar{\alpha}_t=\prod_{s=1}^{t}\alpha_s$. \\

$\sigma_t$ &
Standard deviation of the EI-DDPM reverse transition at step $t$. \\

$\Delta t$ &
Numerical integration interval used during EI-FM sampling. \\

\bottomrule
\caption{Summary of the notations.}
\end{longtable}

\section{An Introduction to DDPM and FM}\label{sec:introDDPMFM}

Here we provide a brief introduction to the conditional versions of DDPM and FM.

\paragraph{DDPM:}
DDPM learns to reverse a forward noising process and generate data by applying the learned reverse process to map samples from a Gaussian distribution $q_0=\mathcal{N}(\bm{0},\bm{I})$ to the target distribution $q_1$. 
In the forward process, a sample starting from $x_0\sim q_1$ is gradually corrupted:
\begin{equation}
    q(x_t|x_{t-1}) = \mathcal{N}(x_{t},\sqrt{1-\beta_t}x_{t-1},\beta_t\bm{I}),\quad t = 1,\cdots, T,
\end{equation}
in which $T$ is the number of total steps and $\beta_1,\cdots,\beta_T$ are pre-defined schedules. $x_T\sim \mathcal{N}(\bm{0},\bm{I})$ when $T$ is sufficiently large. Next, with $\alpha_t:=1-\beta_t$ and $\bar{\alpha}_t:=\prod_{s=1}^t\alpha_s$, DDPM models the reverse process as
\begin{equation}
    p_\theta(x_{t-1}|x_t,z)=\mathcal{N}(x_{t-1};\mu_\theta(x_t,t,z), \sigma_t^2\bm{I}),\quad \mu_\theta(x_t,t,z)=\frac{1}{\sqrt{\alpha_t}}\left(x_t-\frac{\beta_t}{\sqrt{1-\bar{\alpha}_t}}\varepsilon_\theta(x_t,t,z)\right)\label{eq:DDPM_reverse_process}
\end{equation}
in which $z$ is the conditional information, which is a function of $y,\mathcal{Y}$ in EIP-II. $\varepsilon_\theta$ is a neural network (NN) parameterized with $\theta$, and $\sigma_t^2$ is the variance in the reverse process derived from the forward process. With the objective 
of minimizing the expected MSE between a noise $\varepsilon\sim\mathcal{N}(\bm{0},\bm{I})$ and the model's prediction, i.e.,
\begin{equation}
   \arg\min_\theta \mathbb{E}_{x_0,z,t,\varepsilon}[\|\varepsilon_\theta(x_t,t,z)-\varepsilon\|^2],\quad x_t=\sqrt{\bar{\alpha}_{t}}\x_0+\sqrt{1-\bar{\alpha}_{t}}\varepsilon,
\end{equation}
DDPM model learns the reverse process in \eqref{eq:DDPM_reverse_process}.

\paragraph{FM:}
FM aims to learn continuous flows between an initial distribution $q_0$ and the target distribution $q_1$ by learning the velocity fields across time. For $d$-dimensional data, define a stochastic process
$x_t=\Psi_t(x_0,x_1)$, where
$\Psi:[0,1]\times\mathbb{R}^d\times\mathbb{R}^d\rightarrow\mathbb{R}^d$
is twice differentiable in space and time and uniformly Lipschitz in time,
with $\Psi_0(x_0,x_1)=x_0$ and $\Psi_1(x_0,x_1)=x_1$.
Here, $x_1\sim q_1(\cdot\mid z)$ and $x_0\sim q_0$ is sampled
independently of $(x_1,z)$. The velocity field is defined via $v^{\Psi}(x,t,z) = \mathbb{E}[\frac{d}{dt}\Psi_t|X_t=x,z]$. FM aims to learn the velocity field with an NN $\varepsilon_\theta(x_t,t,z)$ parameterized by $\theta$. Similarly, $z$ is the conditional information, which stands for a function of $y, \mathcal{Y}$ in EIP-II. FM's objective minimizes the MSE between the $v^{\Psi}(x,t,z)$ and $\varepsilon_\theta(x_t,t,z)$.
Although $v^{\Psi}(x,t,z)$ is intractable since it is an average over all possible trajectories crossing $x$ conditioned on $z$, one can optimize the objective via the following equivalence \cite{lipman2023flow},
\begin{equation}
   \arg \min_{\theta} \int_{0}^{1} \mathbb{E}[\| \varepsilon_\theta(x_t, t,z) - v^{\Psi}(x_t, t,z)\|^2] dt
    = \arg \min_{\theta} \int_{0}^{1} \mathbb{E}[\| \varepsilon_\theta(x_t, t, z) - \frac{d}{dt} \Psi_t(x_0, x_1)\|^2] dt, \label{eq:FMobj}
\end{equation}
where we recall $x_t = \Psi_t(x_0, x_1)$. Note that $\Psi_t(x_0, x_1)$ can be picked by the user. One common and simple choice is the linear interpolant $\Psi_t(x_0, x_1) = tx_1+(1-t)x_0$, with $\frac{d}{dt} \Psi_t(x_0, x_1)=x_1-x_0$, leading to a concrete objective in \eqref{eq:FMobj} that can be efficiently estimated via Monte-Carlo.

\section{Experiment Details}\label{sec:expdetail}

\subsection{Model Configuration and Computational Cost}
All experiments are run on an NVIDIA L40 GPU with $ 46$ GB memory. The configuration for each experiment is described as follows.

\paragraph{$2$-D Gaussian EIP:}$\phi_w:\mathbb{R}^{4000\times2}\rightarrow\mathbb{R}^{3}$ is implemented according to \citet{settransformer} and consists of an encoder using a single Induced Set Attention Block (ISAB) encoder to capture set-element interactions with linear-time attention via trainable inducing points, and a decoder that performs Pooling by Multihead Attention (PMA), followed by a Set Attention Block (SAB) to model correlations among the pooled outputs, and a final linear projection to the $3$-D ensemble information. Specifically, ISAB, which uses multihead attention with $4$ heads, takes an unordered set $\mathcal{Y}\in \mathbb{R}^{4000\times 2}$ as the input and maps the input to $128$-D embeddings. PMA and SAB both apply multihead attention with $4$ heads and have embedding sizes of $128$. The final linear projection is a linear layer mapping from $128$-D embeddings to $3$-D ensemble information.

$\varepsilon_\theta$ for EI-DDPM and EI-FM consists of Multi-Layer Perceptrons (MLPs), incorporating a time embedding. The network first takes the concatenation of intermediate data $x_t\in \mathbb{R}^2$, the single observation $y\in \mathbb{R}^2$, and the ensemble information $\phi_w(\mathcal{Y})\in \mathbb{R}^3$ as the input and processes it through a $64$-unit hidden layer. The outputs are added with a learned time embedding with time $t$ as an input, and then processed through $64$-unit hidden layers.  Skip connections are employed between the input and output of the main block. The final outputs are $2$-D variables representing the predicted noise / velocity field at time $t$ for EI-DDPM / EI-FM. EI-DDPM has a total number of steps $T=500$. The noise schedule is defined linearly from an initial noise level of $\beta_1 = 1\times10^{-4}$ to a final noise level of $\beta_T = 0.02$ across timesteps $t=1,\dots,T$.  The discretization interval for EI-FM during inference time is set as $\Delta t=0.01$.

\paragraph{Image inversion of MNIST Digits Mixture:} The structures of $\varepsilon_\theta$ and $\phi_w$ are modified to facilitate processing images in this case.
For a set of images, $\phi_w:\mathbb{R}^{128\times28\times28}\rightarrow \mathbb{R}^{28\times 28}$ first processes each image in $\mathcal{Y}$ with a four-stage convolutional encoder with $3\times 3$ convolution kernels for image feature representations. The representation for each image is flattened into $128$-dim variables. Then the representation set is mapped into the ensemble information $\phi_w(\mathcal{Y})\in\mathbb{R}^{28\times 28}$ via a set transformer with the same structure as in $2$-D Gaussian EIP (input and output dimensions adapted).

$\varepsilon_\theta$ employs a U-net structure \citep{ronneberger2015u}, which accepts a matrix of $3$ channels and time $t$ as inputs. The $3$ channels in the matrix are $x_t\in\mathbb{R}^{28\times 28}, y\in\mathbb{R}^{28\times 28}$ and $\phi_w(\mathcal{Y})\in\mathbb{R}^{28\times 28}$. The final outputs are $\mathbb{R}^{28\times 28}$ variables representing the predicted noise / velocity field at time $t$ for EI-DDPM / EI-FM. The total time steps for EI-DDPM is set as $T=500$, and the noise schedule for EI-DDPM remains the same as in $2$-D Gaussian EIP.  The discretization interval for EI-FM during inference time is set as $\Delta t=0.002$.

\paragraph{Particle Physics Data Unfolding:} $\phi_w:\mathbb{R}^{2000\times7}\rightarrow\mathbb{R}^{6}$ shares the same structure as in $2$-D Gaussian EIP, with input and output dimensions adapted. $\varepsilon_\theta$ also shares similar structures as in $2$-D Gaussian EIP, with the number of units in hidden layers changed. The input of the concatenation of intermediate data $x_t\in \mathbb{R}^7$, the single observation $y\in \mathbb{R}^7$, and the ensemble information $\phi_w(\mathcal{Y})\in \mathbb{R}^6$ first passes through a $256$-unit hidden layer and then added with a learned time embedding. The subsequent layers for mapping into $7$-D noise / velocity field consist of $256$-unit and $512$-unit linear layers. The total time steps for EI-DDPM is set as $T=500$, and noise schedule for EI-DDPM remains the same as in $2$-D Gaussian EIP.  The discretization interval for EI-FM during inference time is set as $\Delta t=0.002$.

\paragraph{Seismic Full Waveform Inversion:} $\phi_w:\mathbb{R}^{64\times5\times128\times32}\rightarrow \mathbb{R}^{32\times 32}$ first processes the seismic data in $\mathcal{Y}$ with a four-stage convolutional encoder with $3\times 3$ convolution kernels for feature representations. The representations are flattened into $128$-dim variables. Then the representation set is mapped into the ensemble information $\phi_w(\mathcal{Y})\in\mathbb{R}^{32\times 32}$ via a set transformer with the same structure as in $2$-D Gaussian EIP (input and output dimensions adapted). 

$\varepsilon_\theta$ employs a U-net structure, which accepts $x_t\in\mathbb{R}^{32\times32}$, the seismic data $y\in \mathbb{R}^{5\times128\times32}$, the ensemble information $\phi_w(\mathcal{Y})\in\mathbb{R}^{32\times 32}$ and time $t$ as inputs. The final outputs are $\mathbb{R}^{32\times 32}$ variables representing the predicted noise / velocity field at time $t$ for EI-DDPM / EI-FM. The total time steps for EI-DDPM is set as $T=1000$, and the noise schedule for EI-DDPM remains the same as in $2$-D Gaussian EIP.  The discretization interval for EI-FM during inference time is set as $\Delta t=0.002$.

In terms of runtime, training the proposed ensemble inverse generative models is approximately twice as expensive as training the corresponding vanilla conditional DDPM/FM models, due to the additional cost of learning the ensemble representation. However, this overhead is incurred only offline during training. At inference time, the ensemble representation is computed once for each observation set and then reused for all posterior samples, so the sampling time remains nearly unchanged compared with standard DDPM/FM sampling. Moreover, the proposed approach is substantially faster at inference than iterative optimization or reweighting-based methods, such as OmniFold; in the unfolding experiment, this corresponds to a difference on the order of seconds versus several minutes.

We further verify the practical similarity between the proposed method and its vanilla counterpart in terms of GPU memory usage. Specifically, we select a subset from the FWI task and perform sampling with both EI-FM and vanilla FM, using \texttt{torch.cuda.max\_memory\_allocated()} to measure the peak allocated GPU memory. As reported in Table~\ref{tab:runtime_memory}, the first two stages correspond to memory costs required by both methods, namely loading the data subset onto the GPU and initializing the model. The peak memory usage increases from $792.41$ MB for FM to $1004.41$ MB for EI-FM, corresponding to an additional cost of approximately $26.8\%$. This indicates a moderate overhead rather than a prohibitive one. Since inference efficiency is typically the more practically relevant computational consideration in inverse problems, the proposed method does not introduce a prohibitive computational burden.

\begin{table}[t]
\centering
\begin{tabular}{lcc}
\toprule
\textbf{Stage} & \textbf{EI-FM} & \textbf{Vanilla FM} \\
\midrule
After loading the subset & $246.38$ MB & $246.38$ MB \\
After initializing and loading model & $548.10$ MB & $523.92$ MB \\
After sampling & $1004.41$ MB & $792.41$ MB \\
\bottomrule
\end{tabular}
\caption{GPU memory usage of EI-FM and vanilla FM during sampling on a subset of the FWI task.}\label{tab:runtime_memory}
\vspace{-10pt}
\end{table}

\subsection{Effect of $N,N'$ in $2$-D Gaussian EIP}\label{sec:appNN}

Note that we assume $N'\gg N$ in most cases, i.e., the number of available samples is sufficient to form observation sets that can contain the ensemble information. Fixing the observation set size $N$ for training can contribute to a simpler training pipeline and a more stable optimization process. And this does not impact the inference since size $N$ observation sets are available. However, a fixed $N$ for training is not strictly required. If the number of available observations for inference stays close to $N$, and yet is not fixed, we recommend that users employ random $N$s within a range aligning with the inference requirements during training. In this way, the inference algorithm can automatically work for changeable set sizes, as in the training range.

\begin{figure}[thb]
    \centering
        \begin{tabular}{ccc}
        \begin{subfigure}{0.45\textwidth}
            \includegraphics[width=\linewidth]{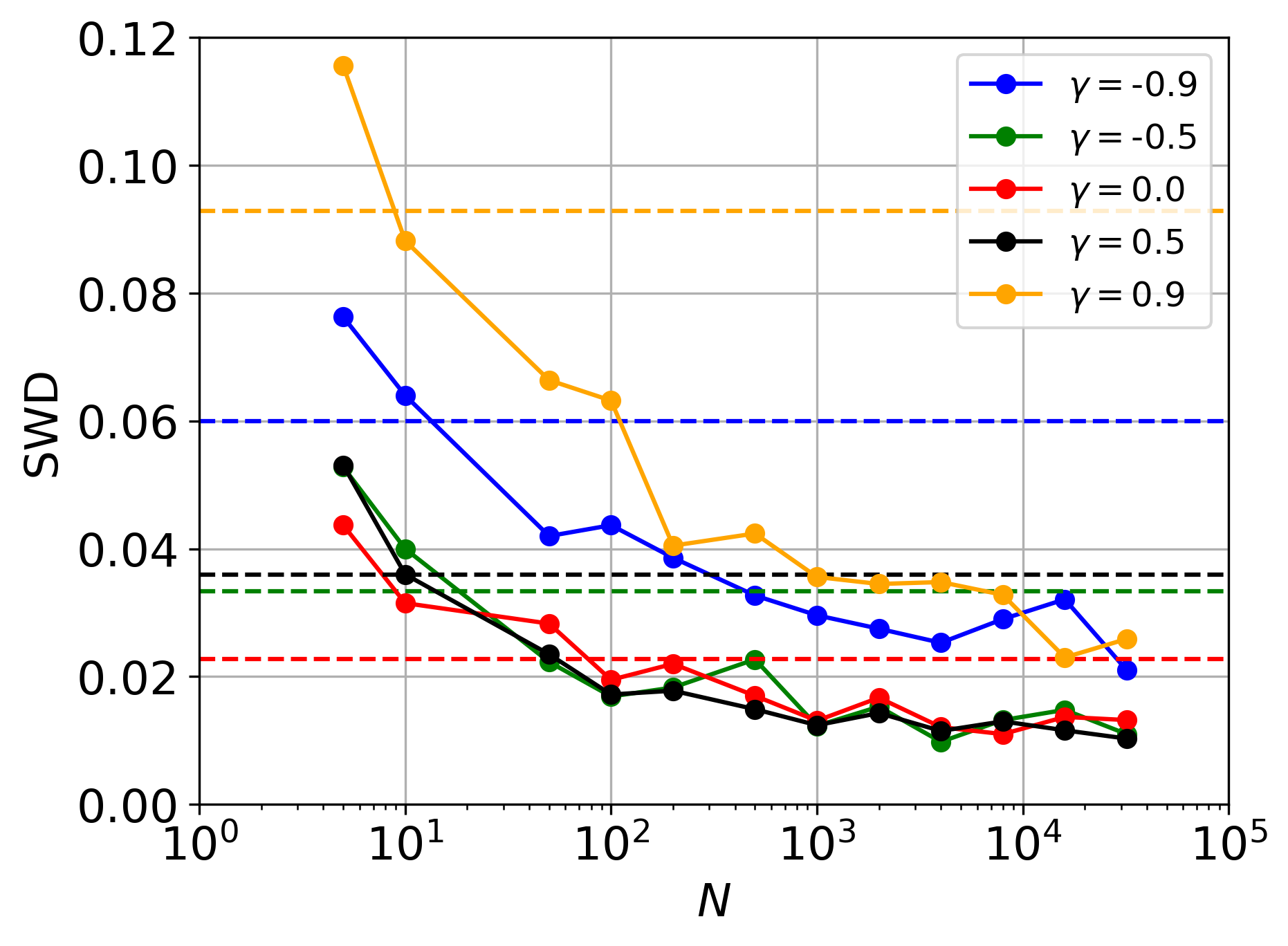}
            \caption{}
            \label{fig:nres}
        \end{subfigure} &\hspace{-15pt}
        \begin{subfigure}{0.45\textwidth}
            \includegraphics[width=\linewidth]{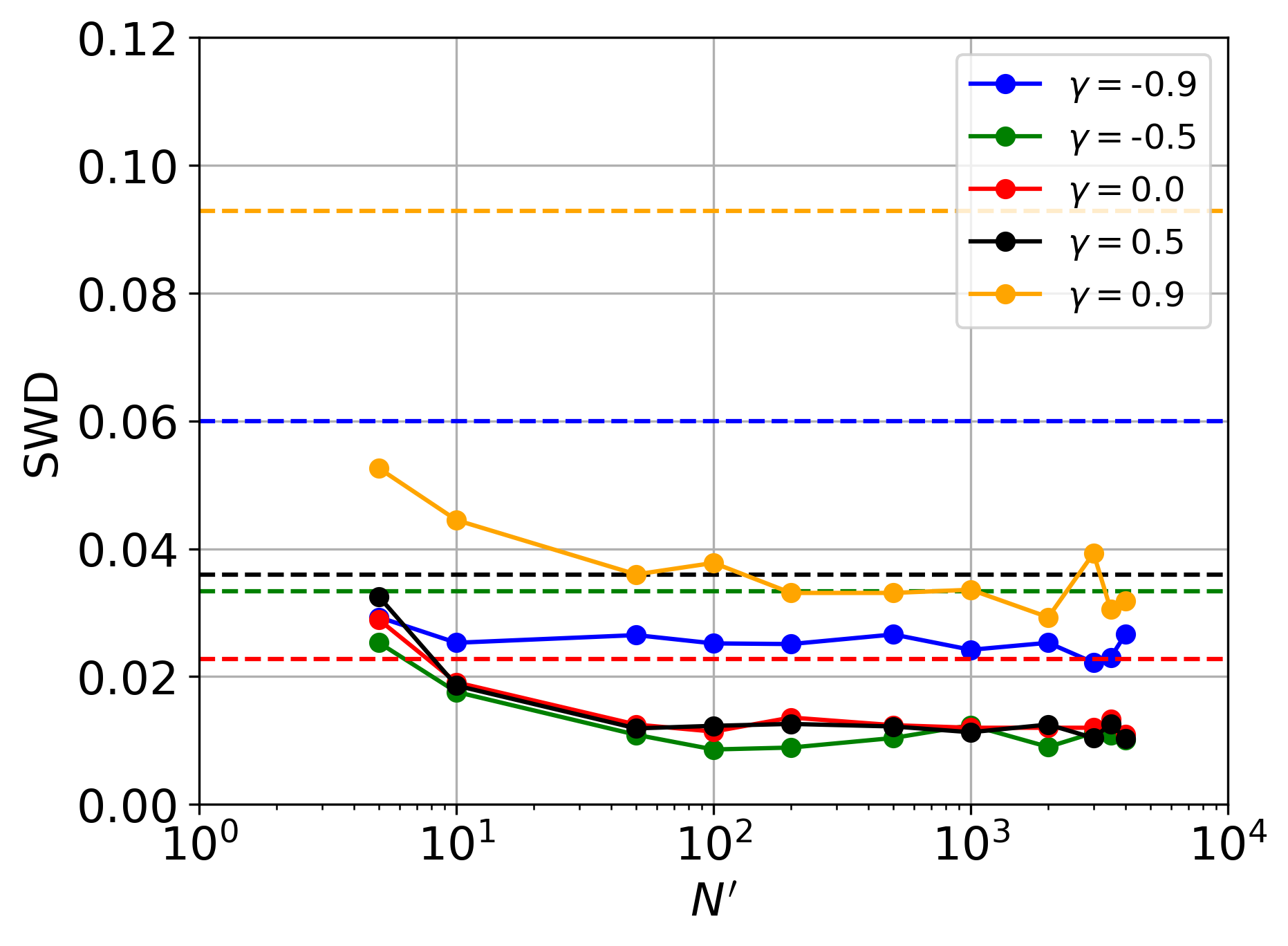}
            \caption{}
            \label{fig:npres}
        \end{subfigure} &\hspace{-15pt}
         \end{tabular}
\caption{Average SWD between the truth and the recovery vs. $N$ and $N'$. The horizontal dashed lines represent the performance of cFM baselines. (a) is for EI-FM trained with different $N$, evaluated over $40000$ samples. We also provide the stats of cFM as a baseline. (b) is for EI-FM trained with $N=4000$, evaluated over $40000$ samples. It is assumed that only $N'$ samples are available during the inference time and Alg.~\ref{alg:sample_imp} is implemented via the duplication strategy.}
\vspace{-10pt}
    \label{fig:NNp}
\end{figure}

Next we numerically investigate the effect of $N$ and $N'$ under the setting of $2$-D Gaussian inverse problem in Sec.~\ref{sec:gaussian}. 
First for the effect of $N$, we train EI-FM with observation set size $N$ from $5$ to $32000$. The results in Fig.~\ref{fig:nres} show that for small $N\leq 10$, the recovery performance is even worse than the baseline cFM without any group information. $\mathcal{Y}$ with too small set sizes cannot represent the ensemble information and even mislead the model in both training and inference. As $N$ grows larger, EI-FM displays its advantage over cFM by leveraging valid ensemble information from $\mathcal{Y}$. The recovery performance evaluated by SWD improves with the growth of 
$N$ and stabilizes when $N$ reaches a sufficiently large value.

Next we consider the cases such that the number of samples to recover $N'$ is smaller than $N$. Take an EI-FM model trained with $N=4000$. Assuming that only $N'$ samples are available during inference time, these $N'$ samples are duplicated until the set has $N$ samples to perform Alg.~\ref{alg:sample_imp}. To evaluate the SWD metric, this process is repeated several times until $40000$ samples are recovered. The results shown in Fig.~\ref{fig:npres} indicate that SWD decreases as $N'$ increases to $4000$. For $N'$ that are not significantly less than $N$, such as $N'=1000$, the duplication strategy can still yield an SWD close to the $N'=4000$ case, since the sets after duplication can still effectively represent the ensemble information. Notably, even with $N'$ as small as $10$, EI-FM slightly outperforms cFM, which performs a sample-wise recovery. This highlights the effect of ensemble information in EIP-II.

\begin{figure}[h]
    \centering
    
    \begin{subfigure}{0.30\textwidth}
        \centering
        \includegraphics[width=\linewidth]{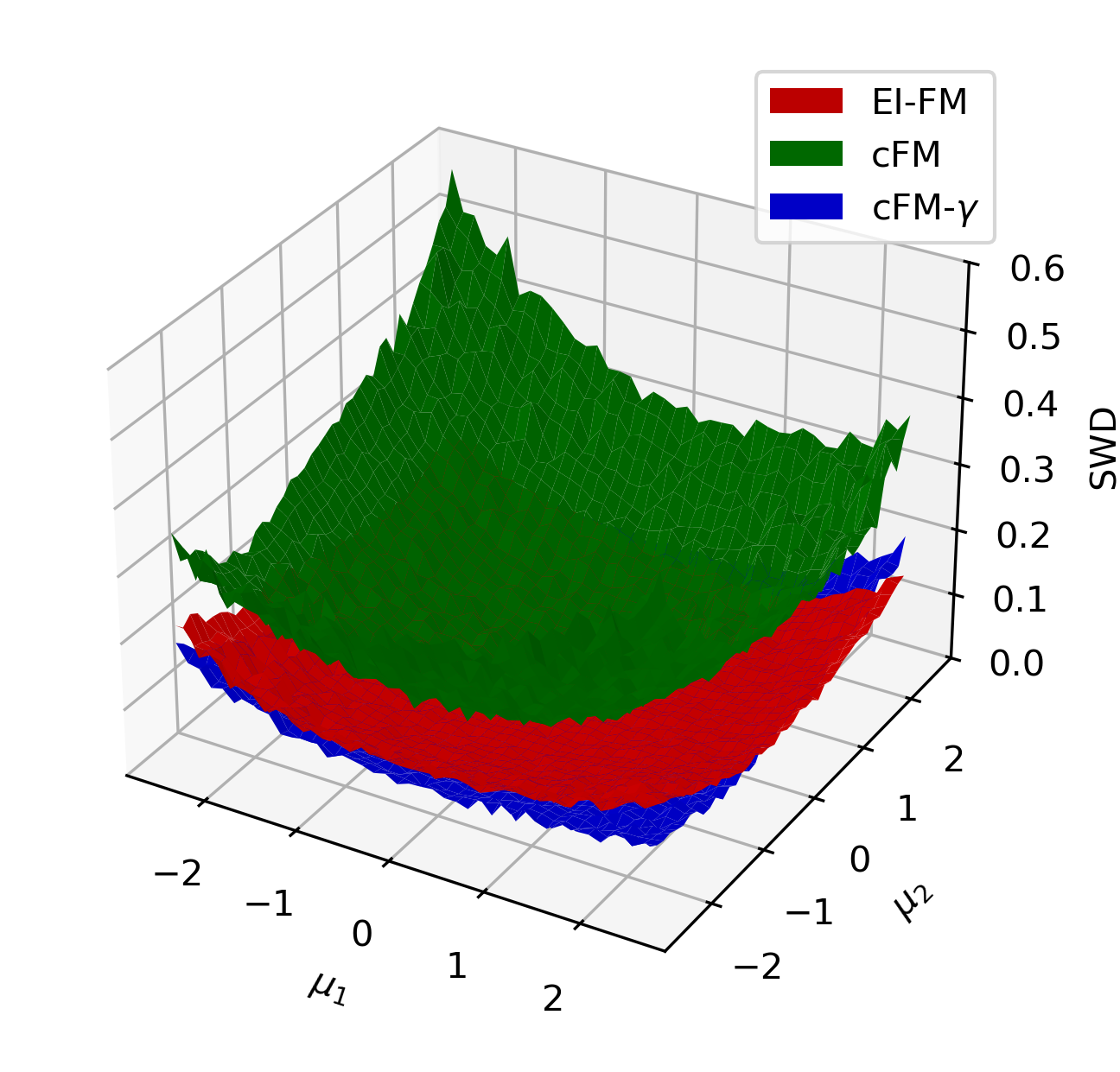}
        \caption{$\gamma_1=-0.9$}
    \end{subfigure}%
    \hfill
    \begin{subfigure}{0.30\textwidth}
        \centering
        \includegraphics[width=\linewidth]{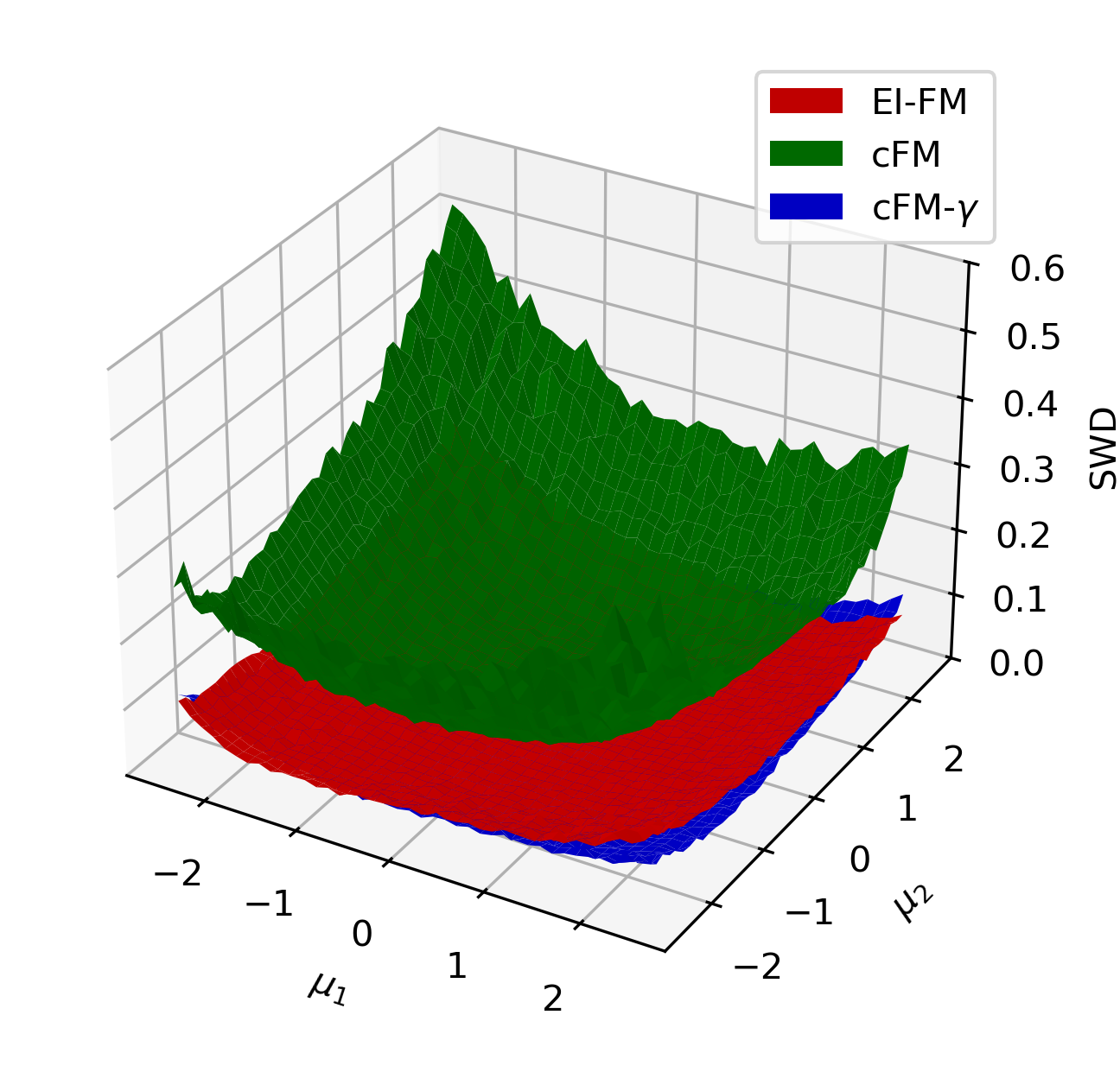}
        \caption{$\gamma_1=-0.5$}
    \end{subfigure}%
    \hfill
    \begin{subfigure}{0.30\textwidth}
        \centering
        \includegraphics[width=\linewidth]{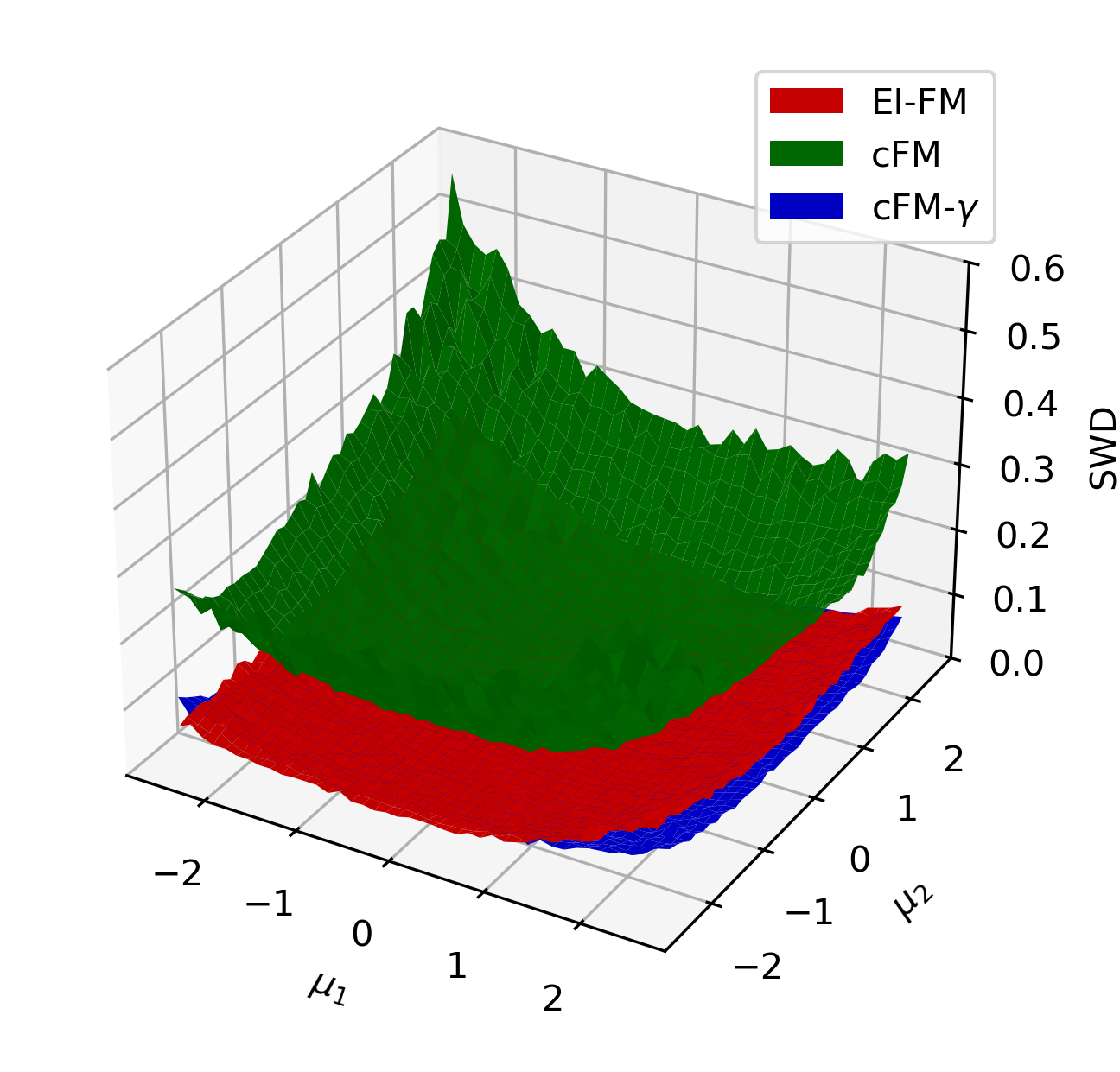}
        \caption{$\gamma_1=0$}
    \end{subfigure}
    
    
    \begin{subfigure}{0.30\textwidth}
        \centering
        \includegraphics[width=\linewidth]{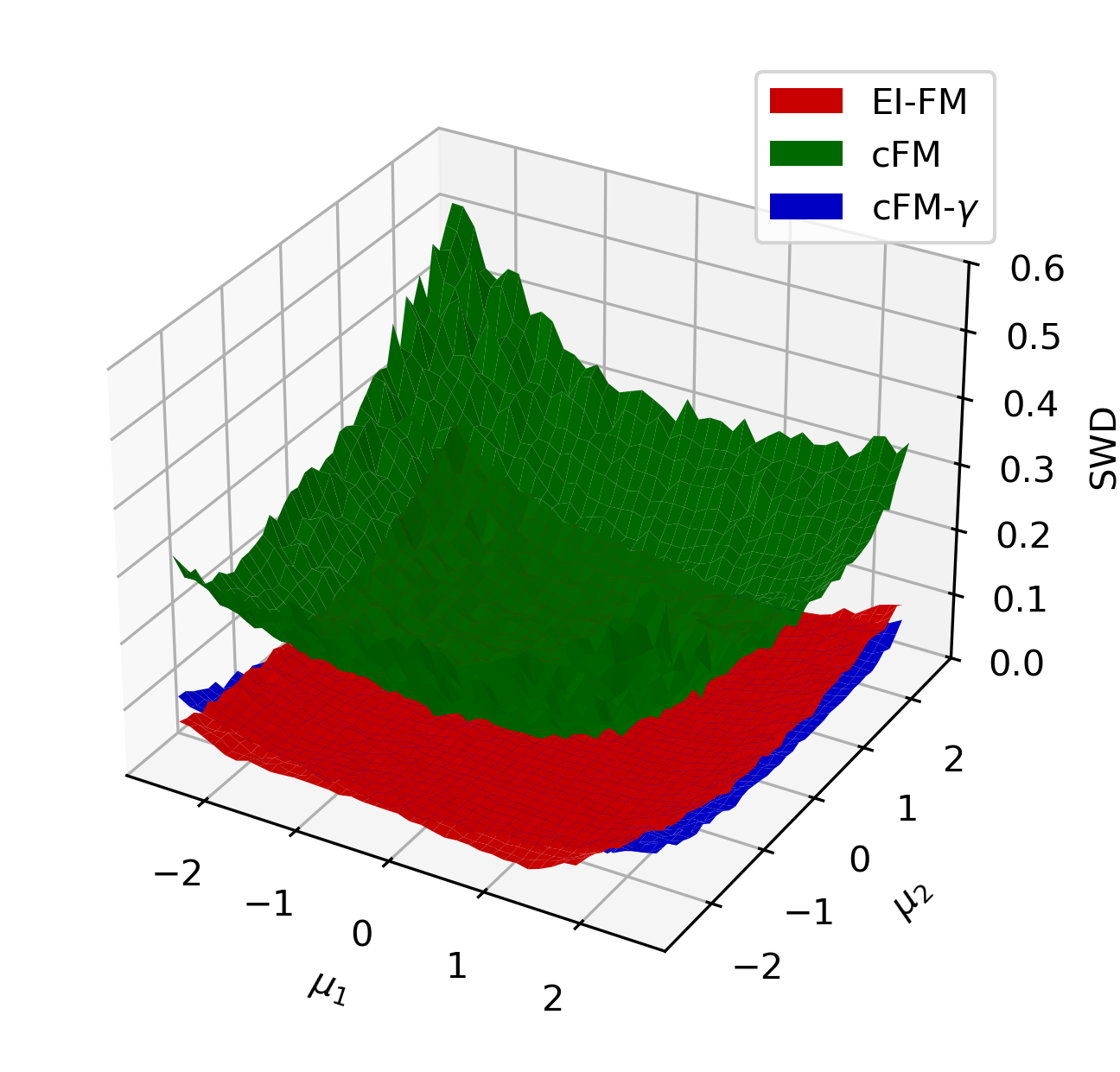}
        \caption{$\gamma_1=0.5$}
    \end{subfigure}%
    \hspace{20pt}
    \begin{subfigure}{0.30\textwidth}
        \centering
        \includegraphics[width=\linewidth]{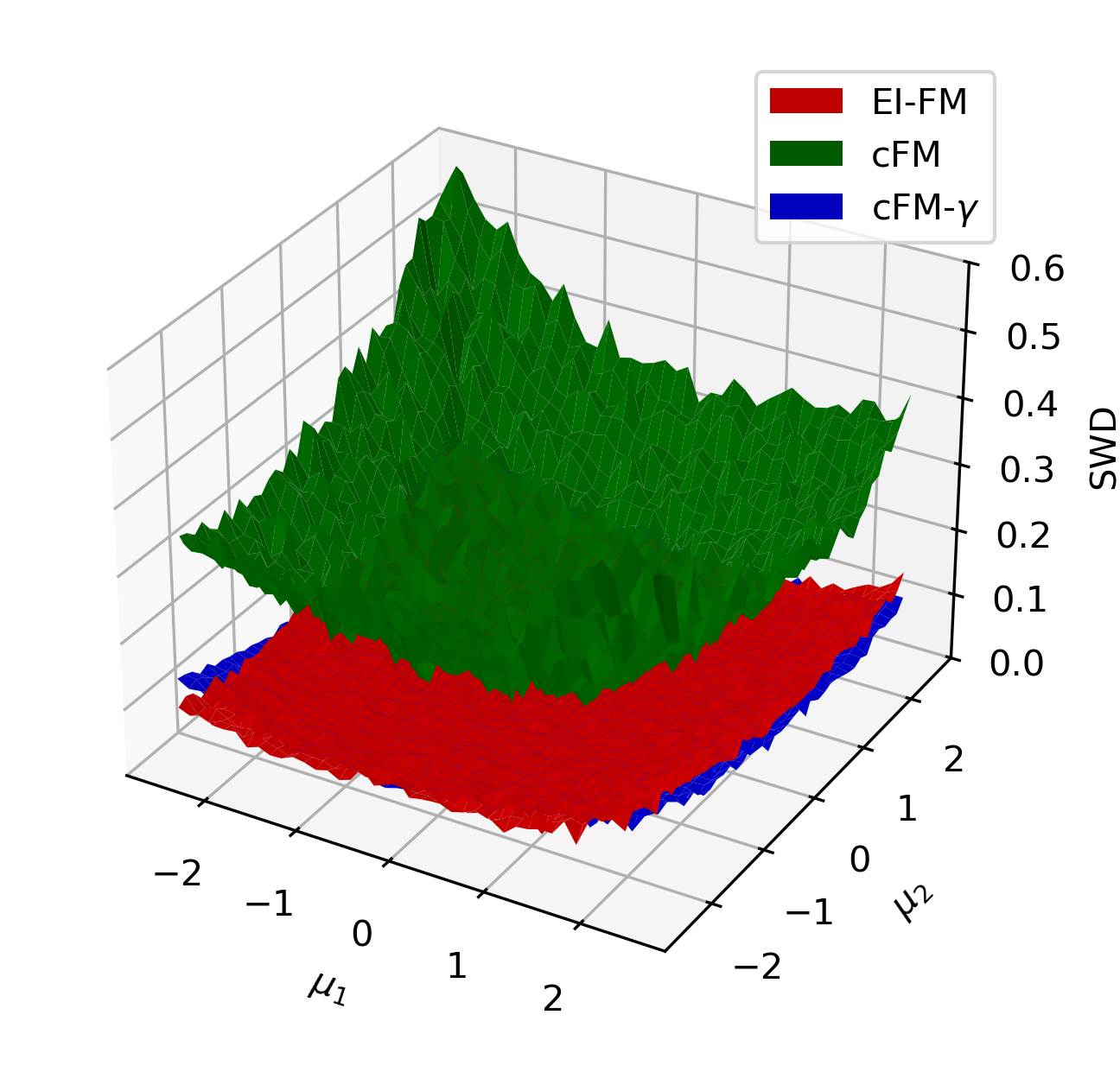}
        \caption{$\gamma_1=0.9$}
    \end{subfigure}
    
    \caption{Average SWD($\downarrow$) between the truth and the recovery vs. $(\mu_1,\mu_2)$ for $\gamma_1=\{-0.9,-0.5,0,0.5,0.9\}$, evaluated over $40000$ samples.}
    \vspace{-10pt}
    \label{fig:GEIP-5}
\end{figure}


\subsection{Extension of $2$-D Gaussian EIP}
Here we present an extension of the $2$-D Gaussian EIP, in which the number of parameters determining the prior increases from $1$ to $3$. Consider the prior
\begin{equation}
    x|\gamma\sim \mathcal{N}\left(\left[\begin{matrix}
        \mu_1\\\mu_2
    \end{matrix}\right], \left[\begin{matrix}
    1&\gamma_1\\\gamma_1&1
\end{matrix}\right]\right),\quad \gamma=(\mu_1,\mu_2,\gamma_1),
\end{equation}
in which $\mu_1,\mu_2,\gamma_1$ are $3$ independent parameters.
Samples from this prior undergo the same forward process as \eqref{eq:toyforward}. One aims to recover the prior given its observation set $\mathcal{Y}$ corresponding to an unknown $\gamma$.

During the training stage, truth-observation pairs resulting from priors with $\gamma_1\in[-0.75,-0.25]\cup[0.25,0.75]$ and $\mu_1,\mu_2\in[-1.5,-0.5]\cup[0.5,1.5]$ are provided. At inference time, we evaluate the recovery performance for priors with $\gamma_1 \in [-1,1]$ and $\mu_1,\mu_2\in [-2,2]$ perturbed by \eqref{eq:toyforward}. We compare EI-FM with \mbox{$\phi_w:\mathbb{R}^{4000\times2}\rightarrow\mathbb{R}^5$}, cFM without any ensemble information and cFM-$\gamma$, which is directly provided with $\gamma=(\mu_1,\mu_2,\gamma_1)$. To illustrate the recovery performance vs. $3$ parameters, we make $3$-D figures, in which x,y axes stand for $\mu_1,\mu_2$ respectively, and each figure corresponds to a specific $\gamma_1$. The $z$ axis stands for the metric of measuring the distribution similarity, i.e., SWD. The results in Fig.~\ref{fig:GEIP-5} show that EI-FM can achieve comparable performance to 
cFM-$\gamma$ across all ranges of $\gamma=(\mu_1,\mu_2,\gamma_1)$ and outperform cFM. EI-FM's close performance to cFM-$\gamma$ (with direct knowledge of the prior) further illustrates that EI-FM can still extract valid ensemble information for posterior inference and generalization as the number of parameters determining the prior increases.

\subsection{Detailed Results of Particle Physics Data Unfolding}\label{sec:unfoldingfull}

In this section, we present the complete results of
    the $1$-D Wasserstein distance between the truth-level data and detector-level data / recovered data via various methods for all $7$ components in the physics process in Table~\ref{tb:physicsresapp}. The detector-level distortion for $\eta$ and $\phi$ is small, and their detector-level distributions have already come close to the true prior. Therefore, some best performances for $\eta,\phi$ appear in the detector-level data, i.e., before unfolding.

\begin{table}[H]
\resizebox{\textwidth}{!}{
\begin{tabular}{|c|c|c|c|c|c|c|c|c|c|c|c|c|}
\hline
WD ($\downarrow$) & Name & Detector & EI-DDPM & EI-FM & cDDPM & cFM & GDDPM-v & GDDPM & \begin{tabular}[c]{@{}c@{}}OmniFold\\ -best\end{tabular} & \begin{tabular}[c]{@{}c@{}}OmniFold\\ -combine\end{tabular} & SBUnfold &Sourcerer \\ \hline
\multirow{4}{*}{$p_T$} & Leptoquark & $31.85$ & $0.44$ & $0.44$ & $1.08$ & $2.65$ & $0.73$ & $0.44$ & $\textcolor{red}{\mathbf{0.19}}$ & $0.82$ & $18.10$ &$7.96$ \\ \cline{2-13} 
 & $t\bar{t}$ (CT14lo, Vincia) & $24.18$ & $0.44$ & $\textcolor{red}{\mathbf{0.23}}$ & $1.01$ & $3.36$ & $1.36$ & $0.55$ & $0.60$ & $0.52$ & $1.88$ &$15.14$ \\ \cline{2-13} 
 & $W+$jets (CT14lo) & $18.60$ & $0.60$ & $0.44$ & $2.41$ & $4.14$ & $0.53$ & $0.48$ & $\textcolor{red}{\mathbf{0.11}}$ & $0.37$ & $21.07$ &$25.96$ \\ \cline{2-13} 
 & $Z+$jets (CTEQ6L1) & $15.81$ & $0.51$ & $\textcolor{red}{\mathbf{0.45}}$ & $2.55$ & $5.55$ & $2.25$ & $1.98$ & $0.48$ & $0.64$ & $25.18$ &$19.59$ \\ \hline
 \multirow{4}{*}{$\eta$} 
& Leptoquark& $0.00074$ & $0.00079$ & $0.00096$ & $0.00182$ & $0.00272$ & $0.00255$ & $\textcolor{red}{\mathbf{0.00056}}$ & $0.01936$ & $0.00758$ & $0.04350$ &$0.02079$ \\ \cline{2-13} 
& $t\bar{t}$ (CT14lo, Vincia)& $0.00080$ & $0.00075$ & $0.00095$ & $0.00128$ & $0.00298$ & $0.00363$ & $\textcolor{red}{\mathbf{0.00071}}$ & $0.00689$ & $0.00979$ & $0.03795$ &$0.01997$ \\ \cline{2-13} 
& $W+$jets (CT14lo)& $\textcolor{red}{\mathbf{0.00060}}$ & $0.00096$ & $0.00109$ & $0.00186$ & $0.00406$ & $0.00375$ & $0.00080$ & $0.01945$ & $0.01596$ & $0.06352$ &$0.00728$ \\ \cline{2-13} 
& $Z+$jets (CTEQ6L1)& $\textcolor{red}{\mathbf{0.00065}}$ & $0.00093$ & $0.00111$ & $0.00202$ & $0.00466$ & $0.00298$ & $0.00072$ & $0.00934$ & $0.02069$ & $0.07880$ &$0.03140$ \\ \hline 
\multirow{4}{*}{$\phi$} 
& Leptoquark& $0.00140$ & $0.00091$ & $\textcolor{red}{\mathbf{0.00069}}$ & $0.00662$ & $0.00379$ & $0.00342$ & $0.00142$ & $0.01492$ & $0.00534$ & $0.01452$ &$0.01891$ \\ \cline{2-13} 
& $t\bar{t}$ (CT14lo, Vincia)& $0.00144$ & $0.00096$ & $\textcolor{red}{\mathbf{0.00078}}$ & $0.00718$ & $0.00381$ & $0.00383$ & $0.00158$ & $0.00609$ & $0.00397$ & $0.01493$ &$0.01137$ \\ \cline{2-13} 
& $W+$jets (CT14lo)& $0.00153$ & $0.00092$ & $\textcolor{red}{\mathbf{0.00074}}$ & $0.00803$ & $0.00401$ & $0.00373$ & $0.00159$ & $0.00689$ & $0.00625$ & $0.01426$&$0.01482$  \\ \cline{2-13} 
& $Z+$jets (CTEQ6L1)& $0.00153$ & $0.00107$ & $\textcolor{red}{\mathbf{0.00071}}$ & $0.00836$ & $0.00427$ & $0.00396$ & $0.00177$ & $0.03053$ & $0.00388$ & $0.01552$&$0.03828$  \\ \hline 
\multirow{4}{*}{$E$} & Leptoquark & $83.87$ & $\textcolor{red}{\mathbf{0.46}}$ & $0.76$ & $4.70$ & $2.66$ & $1.47$ & $0.63$ & $1.08$ & $0.47$ & $13.99$ &$57.42$ \\ \cline{2-13} 
 & $t\bar{t}$ (CT14lo, Vincia) & $69.70$ & $0.77$ & $\textcolor{red}{\mathbf{0.66}}$ & $2.96$ & $3.29$ & $1.54$ & $0.89$ & $0.83$ & $1.41$ & $4.83$ &$104.13$ \\ \cline{2-13} 
 & $W+$jets (CT14lo) & $90.42$ & $1.08$ & $1.60$ & $4.56$ & $4.64$ & $3.38$ & $1.60$ & $\textcolor{red}{\mathbf{0.56}}$ & $3.05$ & $23.67$ &$94.95$ \\ \cline{2-13} 
 & $Z+$jets (CTEQ6L1) & $83.18$ & $\textcolor{red}{\mathbf{0.81}}$ & $1.19$ & $6.83$ & $12.62$ & $6.25$ & $7.04$ & $1.44$ & $2.22$ & $40.67$&$79.37$  \\ \hline
\multirow{4}{*}{$p_x$} & Leptoquark & $20.26$ & $\textcolor{red}{\mathbf{0.21}}$ & $0.26$ & $0.95$ & $1.39$ & $0.73$ & $0.25$ & $0.41$ & $0.43$ & $10.53$ &$5.89$ \\ \cline{2-13} 
 & $t\bar{t}$ (CT14lo, Vincia) & $15.40$ & $0.19$ & $\textcolor{red}{\mathbf{0.13}}$ & $0.65$ & $1.03$ & $0.82$ & $0.31$ & $0.41$ & $0.30$ & $1.00$ &$11.22$ \\ \cline{2-13} 
 & $W+$jets (CT14lo) & $11.84$ & $0.26$ & $\textcolor{red}{\mathbf{0.21}}$ & $1.07$ & $0.90$ & $0.75$ & $\textcolor{red}{\mathbf{0.21}}$ & $0.24$ & $0.22$ & $8.75$&$11.04$  \\ \cline{2-13} 
 & $Z+$jets (CTEQ6L1) & $10.06$ & $0.23$ & $\textcolor{red}{\mathbf{0.19}}$ & $1.19$ & $0.66$ & $1.22$ & $1.07$ & $0.35$ & $0.31$ & $16.08$ &$10.74$ \\ \hline
 \multirow{4}{*}{$p_y$} 
& Leptoquark& $20.29$ & $\textcolor{red}{\mathbf{0.25}}$ & $\textcolor{red}{\mathbf{0.25}}$ & $0.95$ & $1.63$ & $0.36$ & $\textcolor{red}{\mathbf{0.25}}$ & $0.53$ & $0.56$ & $10.81$ &$7.41$ \\ \cline{2-13} 
& $t\bar{t}$ (CT14lo, Vincia)& $15.39$ & $0.23$ & $\textcolor{red}{\mathbf{0.13}}$ & $0.90$ & $1.47$ & $0.65$ & $0.31$ & $0.45$ & $0.27$ & $0.89$ &$10.80$ \\ \cline{2-13} 
& $W+$jets (CT14lo)& $11.84$ & $0.28$ & $\textcolor{red}{\mathbf{0.18}}$ & $1.51$ & $1.54$ & $0.28$ & $0.19$ & $0.22$ & $0.26$ & $8.86$ &$14.32$ \\ \cline{2-13} 
& $Z+$jets (CTEQ6L1)& $10.06$ & $0.25$ & $\textcolor{red}{\mathbf{0.19}}$ & $1.87$ & $1.44$ & $1.28$ & $1.09$ & $0.52$ & $0.27$ & $15.72$ &$11.77$ \\ \hline 
\multirow{4}{*}{$p_z$} 
& Leptoquark& $70.72$ & $0.67$ & $\textcolor{red}{\mathbf{0.56}}$ & $6.78$ & $5.04$ & $0.99$ & $0.86$ & $3.15$ & $1.00$ & $17.94$ &$45.64$ \\ \cline{2-13} 
& $t\bar{t}$ (CT14lo, Vincia)& $60.38$ & $0.86$ & $\textcolor{red}{\mathbf{0.52}}$ & $6.22$ & $4.21$ & $0.87$ & $1.06$ & $2.09$ & $2.28$ & $11.53$ &$74.01$ \\ \cline{2-13} 
& $W+$jets (CT14lo)& $84.96$ & $\textcolor{red}{\mathbf{1.18}}$ & $1.41$ & $7.48$ & $5.82$ & $3.83$ & $1.57$ & $4.25$ & $3.52$ & $12.38$ &$70.12$ \\ \cline{2-13} 
& $Z+$jets (CTEQ6L1)& $78.70$ & $\textcolor{red}{\mathbf{1.06}}$ & $1.15$ & $6.50$ & $7.09$ & $5.89$ & $6.89$ & $2.90$ & $2.96$ & $33.05$ &$62.07$ \\ \hline 
\end{tabular}}
\caption{Result of data recovery performance on $4$ unseen physics distributions. We report
    the $1$-D Wasserstein distance between the truth-level data and detector-level data / recovered data via various methods for $p_T,\eta, \phi,E, p_x, p_y, p_z$. The best results are noted in red.}
    \label{tb:physicsresapp}
\end{table}

\vspace{-10pt}

\subsection{TARP Coverage as an Additional Metric for Posterior Accuracy}\label{app:tarp}

To further assess the accuracy of posterior samplers, we employ the Test of Accuracy with Random Points (TARP) expected coverage metric that is introduced in \citet{Tarp}. Let $(x^\star,y)$ denote the truth-observation pairs,
TARP assesses whether an estimated posterior
$\hat{p}(x|y)$ correctly approximates the true posterior ${p}(x|y)$ by examining how often credible regions constructed from $\hat{p}$ contain the truth $x^*$.  For each truth--observation pair $(x^\star,y)$, a reference point $x_r$ is randomly drawn. For a credibility level $1-\alpha$, the TARP region is defined as
\begin{equation}
D_{x_r}(\hat p,\alpha,y,d)
=
\left\{
x:
d(x,x_r)
\leq
R_{x_r}(\hat p,\alpha,y,d)
\right\},
\end{equation}
where $d(\cdot,\cdot)$ is a distance function and the radius $R_{x_r}(\hat p,\alpha,y,d)$ is chosen such that
\begin{equation}
\int_{D_{x_r}(\hat p,\alpha,y,d)}
\hat p(x\mid y)\,dx
=
1-\alpha.
\end{equation}
The expected coverage for credibility level $1-\alpha$ is
\begin{equation}
\operatorname{ECP}(\hat{p},\alpha,D_{x_r})
=
\mathbb{E}_{(x^\star,y),x_r}
\left[
\mathbf{1}
\left\{
x^\star\in D_{x_r}(\hat{p},\alpha,y)
\right\}
\right].
\end{equation}
Under the regularity conditions and sufficiently rich choices of reference-point positions specified in \citet{Tarp}, matching this identity for all $\alpha\in(0,1)$ characterizes the true posterior, i.e., $\hat p(x\mid y)=p(x\mid y)$.
 In practice, one evaluates this metric by scanning over credibility levels $1-\alpha$ and plotting the ECP vs. credibility level curve for examining the posterior's correctness. For each curve, we also let $\alpha$ follow a uniform distribution in $(0,1)$ and report $e=\mathbb{E}_{\alpha\sim\mathcal{U}(0,1)}
\left[
\left|
\operatorname{ECP}(\hat p,\alpha,D_{x_r})-(1-\alpha)
\right|
\right]$ for each estimated posterior $\hat{p}$ to measure its deviation from the ideal case.

\begin{figure}[thb]
    \centering
    \includegraphics[width=0.32\linewidth]{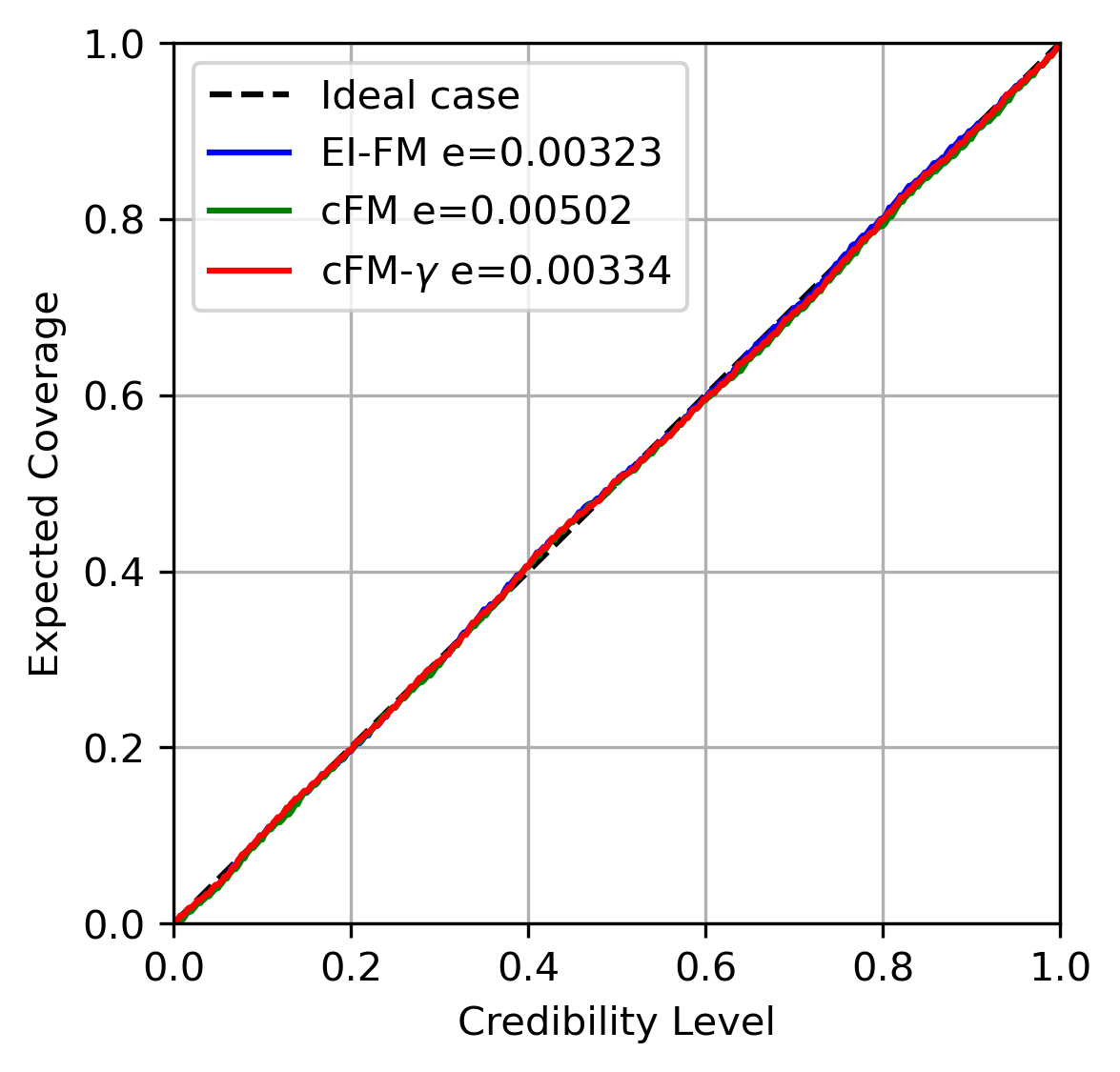}
    \includegraphics[width=0.32\linewidth]{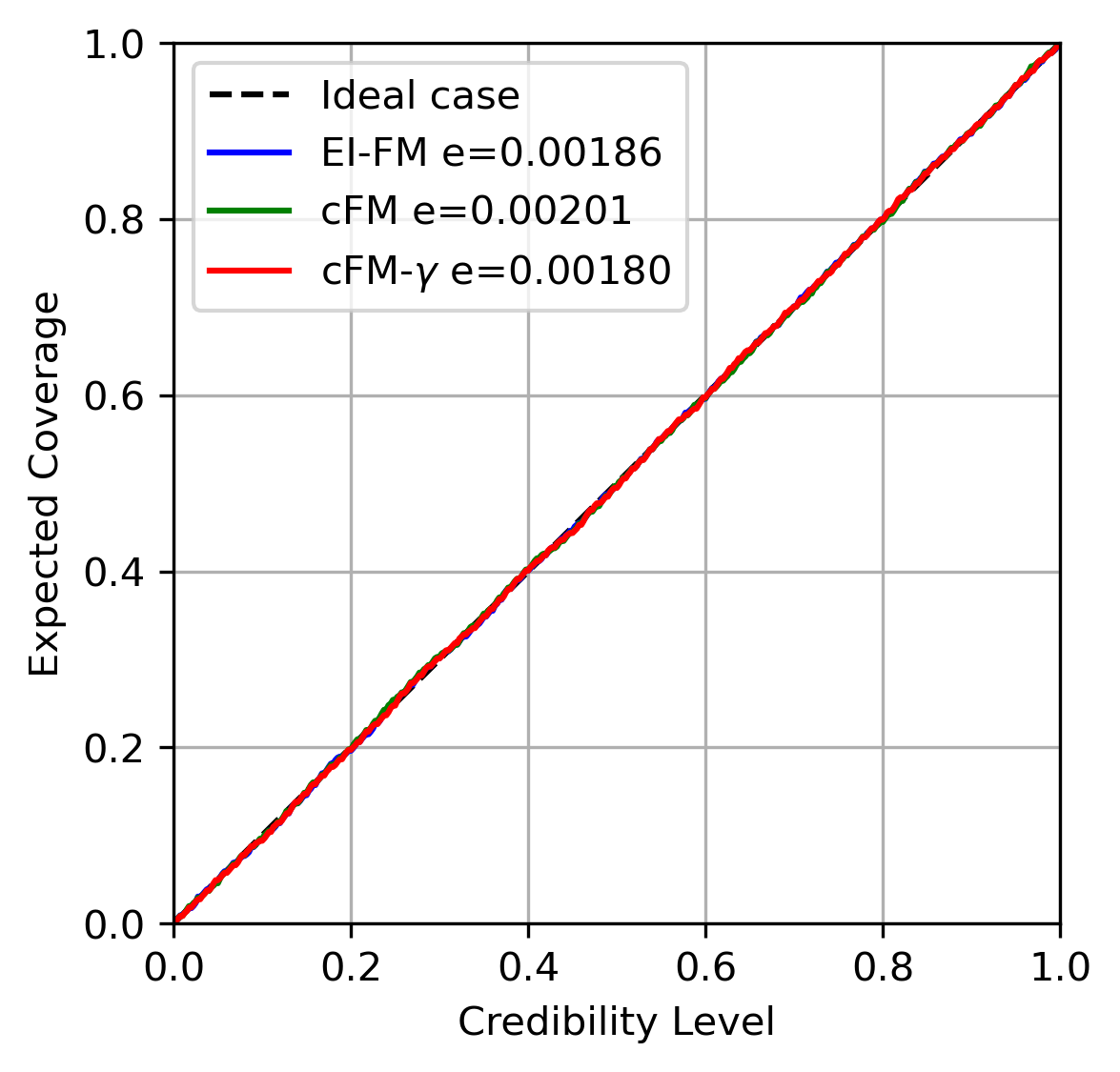}
    \includegraphics[width=0.32\linewidth]{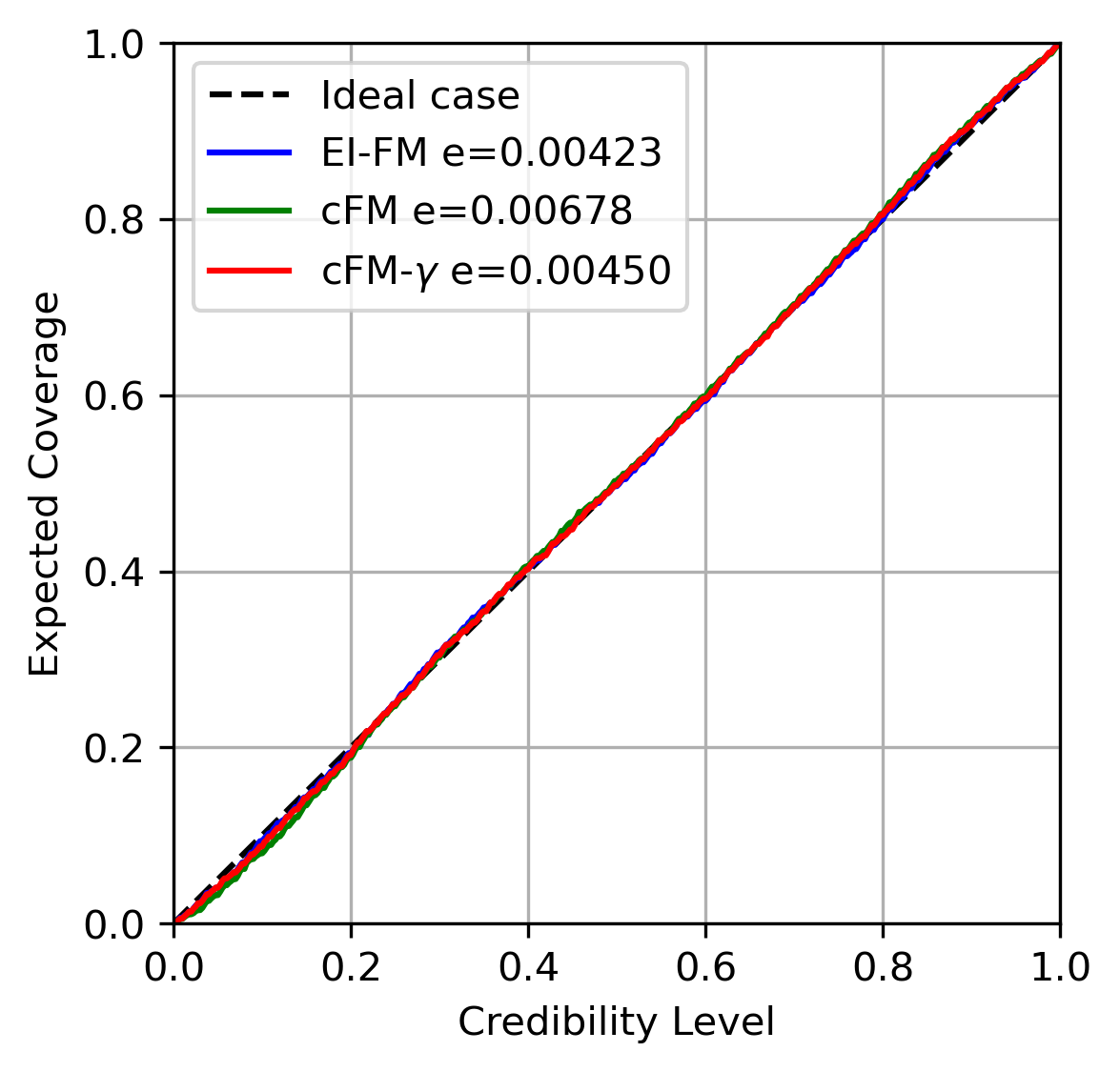}
    \caption{ECP vs. credibility level curves for different models in the $2$-D Gaussian EIP at Sec.~\ref{sec:gaussian}. Left: $\gamma=-0.9$. Middle: $\gamma=0$. Right: $\gamma=0.9$.}
    \label{fig:stntarp}
\end{figure}

\begin{figure}[thb]
    \centering
    \includegraphics[width=0.4\linewidth]{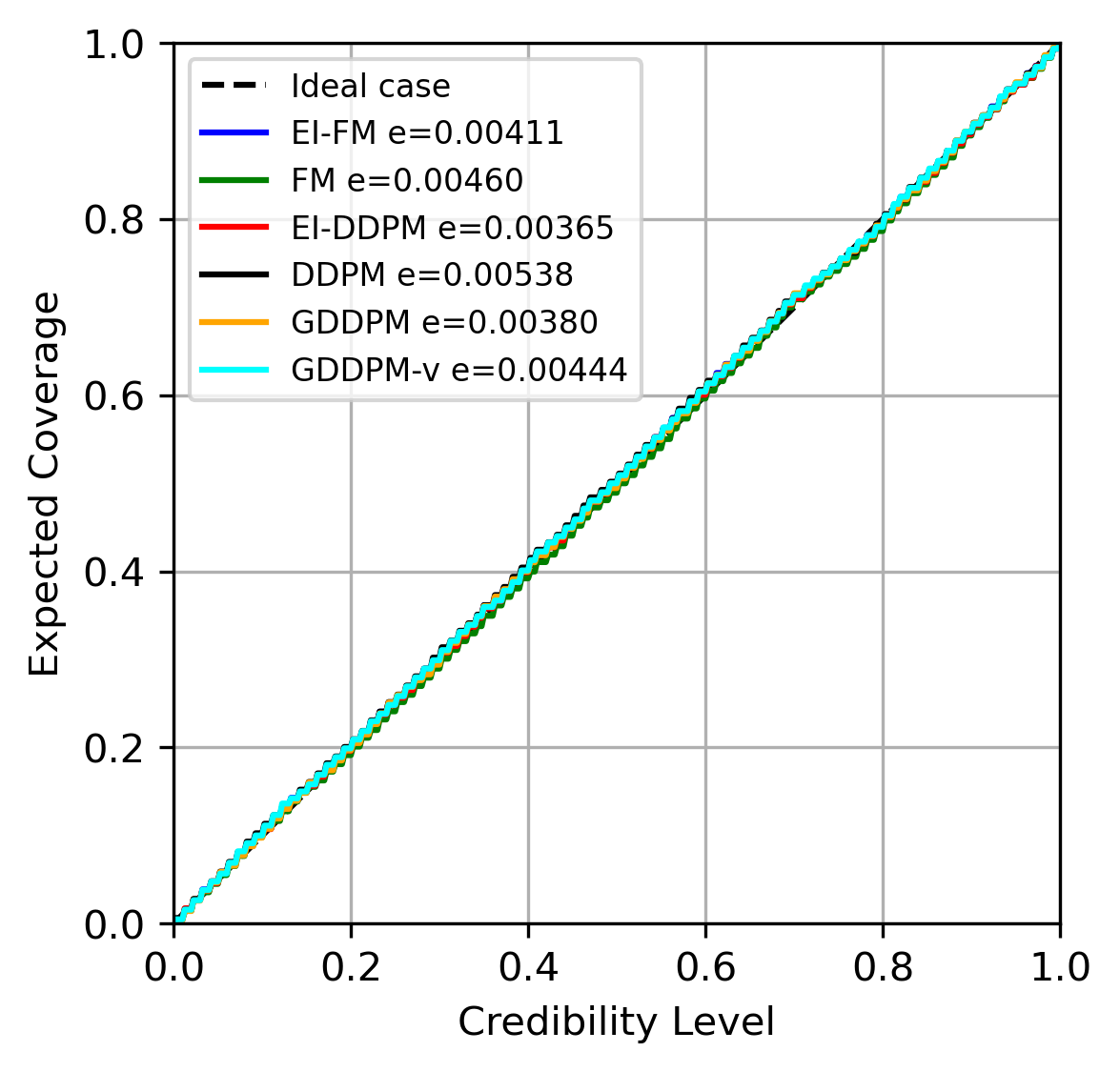}
    \includegraphics[width=0.4\linewidth]{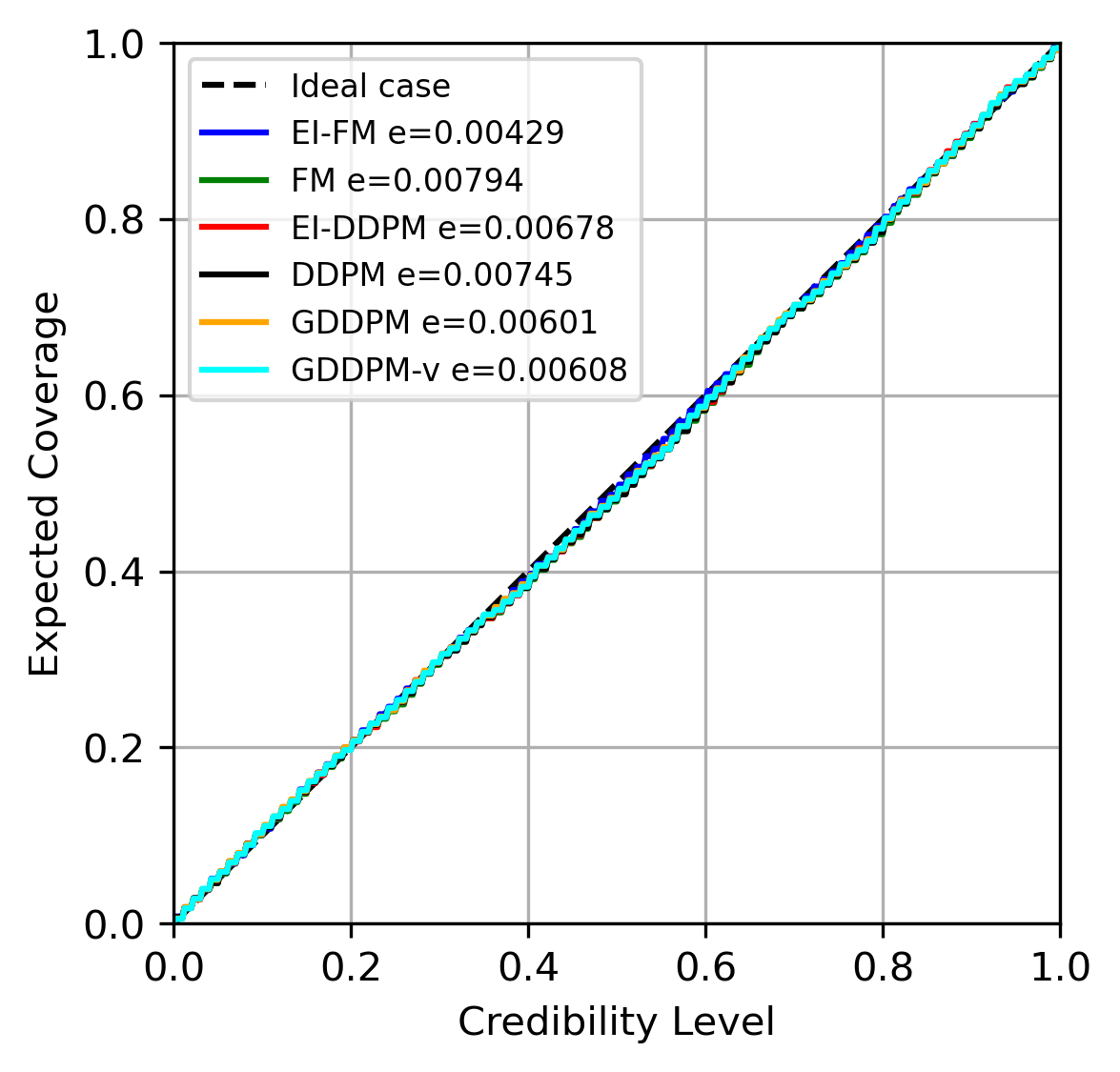}
    \caption{ECP vs. credibility level curves for different models in the particle physics unfolding task at Sec.~\ref{sec:unfold}. Left: Leptoquark process. Right: $t\bar{t}$ (CT14lo, Vincia) process.}
    \label{fig:unfoldtarp}
\end{figure}

The ECP vs. credibility level results for various posterior samplers in the $2$-D Gaussian EIP at Sec.~\ref{sec:gaussian} and in the HEP unfolding at Sec.~\ref{sec:unfold} are shown in Fig.~\ref{fig:stntarp} and Fig.~\ref{fig:unfoldtarp}, respectively. While all posterior samplers' ECP vs. credibility level curves are close to the ideal case, those with ensemble information (e.g., EI-FM, EI-DDPM, cFM-$\gamma$, GDDPMs) display slightly smaller deviations $e$ from the ideal case, illustrating the effectiveness of ensemble information in modeling posteriors. 


\subsection{Visualization of Full Waveform Inversion Results}\label{sec:visFWI}

\begin{figure}[h]
\centering

\renewcommand{\arraystretch}{1.25}

\vspace{-10pt}
\begin{tabular}{>{\RaggedRight\arraybackslash}m{0.15\textwidth} >{\centering\arraybackslash}m{0.58\textwidth}}
&  \\
FlatVel-A      & \includegraphics[width=\linewidth]{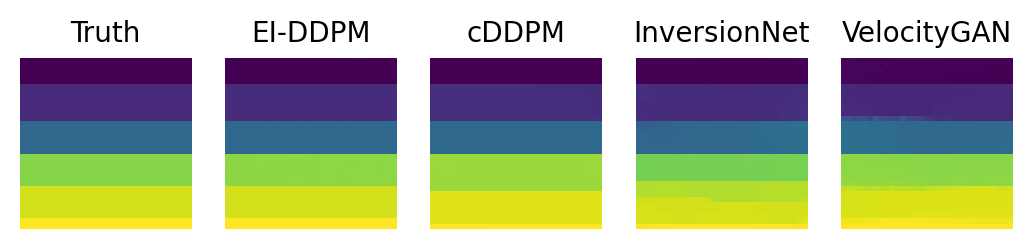} \\
FlatVel-B      & \includegraphics[width=\linewidth]{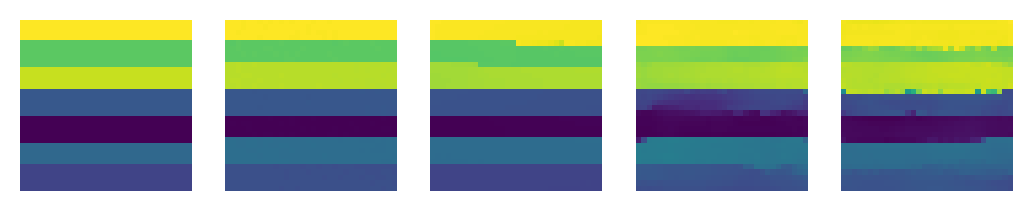} \\
CurveVel-A     & \includegraphics[width=\linewidth]{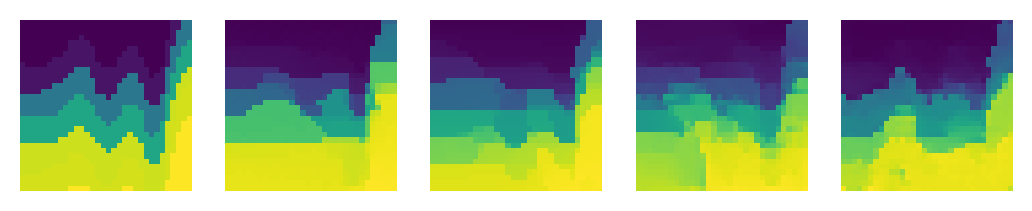} \\
CurveVel-B     & \includegraphics[width=\linewidth]{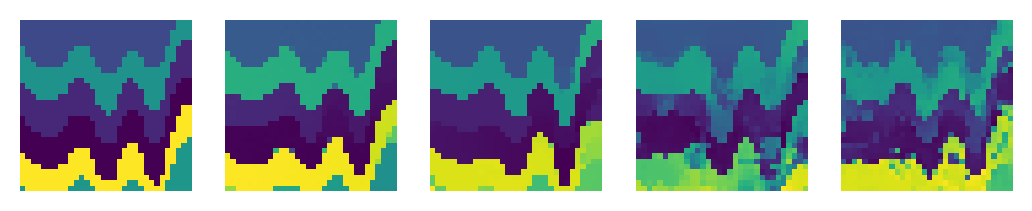} \\
FlatFault-A    & \includegraphics[width=\linewidth]{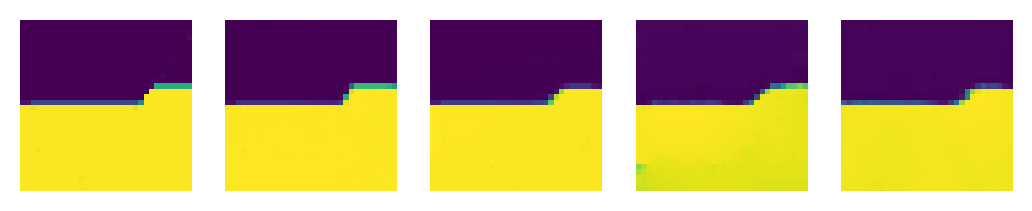} \\
FlatFault-B    & \includegraphics[width=\linewidth]{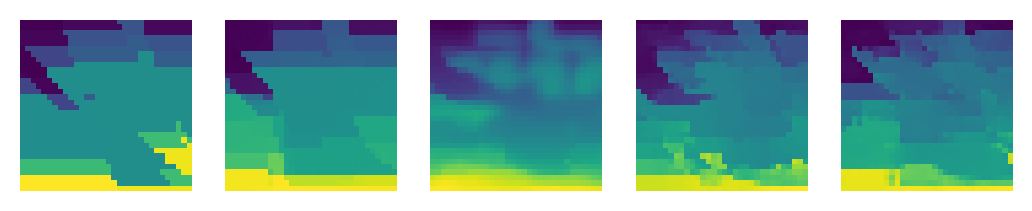} \\
CurveFault-A   & \includegraphics[width=\linewidth]{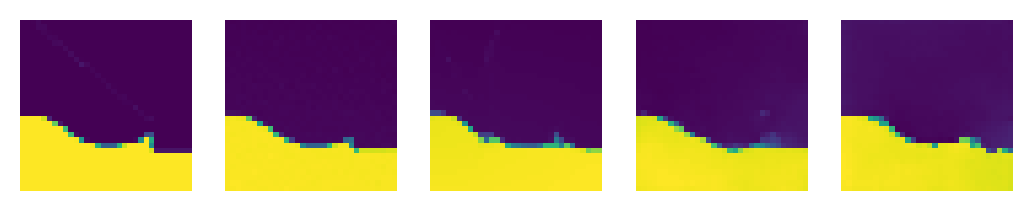} \\
CurveFault-B   & \includegraphics[width=\linewidth]{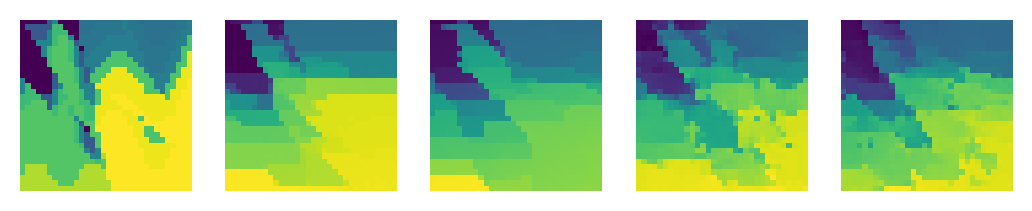} \\
Style-A        & \includegraphics[width=\linewidth]{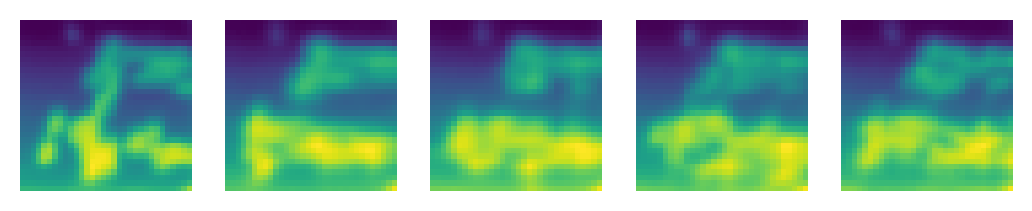} \\
Style-B        & \includegraphics[width=\linewidth]{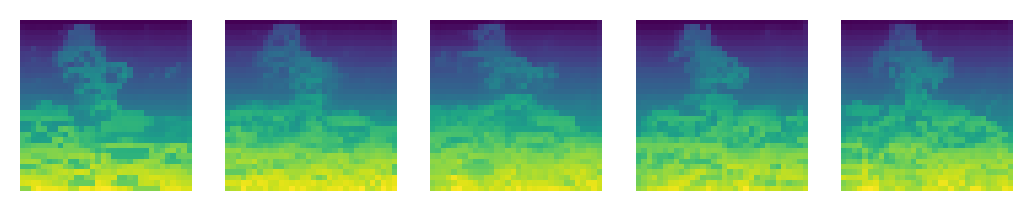} \\
\end{tabular}
\captionsetup{font=small}
\caption{Visualization of FWI results of the compared methods from each family.}
\label{fig:fwi_examples}
\end{figure}

Fig.~\ref{fig:fwi_examples} presents a visualization of the recovered velocity maps from each family obtained via various methods. EI-DDPM preserves the structural patterns of the subsurface maps more faithfully, whereas the baselines tend to introduce stronger structural distortions or miss key geometric features.

\newpage

\end{document}

%% file: math_commands.tex
\usepackage{amsmath,amsfonts,bm}

\def\eqref#1{equation~\ref{#1}}

\def\1{\bm{1}}

\DeclareMathAlphabet{\mathsfit}{\encodingdefault}{\sfdefault}{m}{sl}
\SetMathAlphabet{\mathsfit}{bold}{\encodingdefault}{\sfdefault}{bx}{n}

